%% file: main.tex
\documentclass{article}
\PassOptionsToPackage{numbers}{natbib}
    \usepackage[final]{neurips_2024}

\usepackage[utf8]{inputenc} 
\usepackage[T1]{fontenc}    
\usepackage{hyperref}       
\usepackage{url}            
\usepackage{booktabs}       
\usepackage{amsfonts}       
\usepackage{amsmath}
\usepackage{amssymb} 
\usepackage{amsthm}
\usepackage{stmaryrd}
\usepackage[scaled=0.75]{beramono}
\usepackage{wrapfig}

\usepackage{nicefrac}       
\usepackage{microtype}      
\usepackage{xcolor}         
\usepackage{algorithm}
\usepackage[noend]{algpseudocode}
\usepackage{listings}
\usepackage{mathtools}
\usepackage{multirow}
\usepackage{tcolorbox}

\usepackage{color, soul}
\definecolor{grey}{rgb}{0.9,0.9,0.9}
\usepackage{balance}
\usepackage{textcomp}

\usepackage{listings} 
\usepackage{alltt}
\usepackage{latexsym}
\usepackage{xspace}
\usepackage{url}
\usepackage{paralist}
\usepackage{pdfpages}
\usepackage{listings} 
\usepackage{alltt}
\usepackage{hyperref}
\usepackage{booktabs}  
\usepackage{subcaption} 
\usepackage{url}
\usepackage{hhline}
\usepackage{multicol}
\usepackage{makecell}
\usepackage{todonotes}
\usepackage[font={small}]{caption}
\usepackage[inline,shortlabels]{enumitem}
\usepackage{color}
\usepackage[export]{adjustbox}
\usepackage{natbib}
\usepackage{bbm}

\title{Grammar-Aligned Decoding}

\author{%
\makebox[\textwidth][c]{%
  Kanghee Park$^1\thanks{Equal contribution}\quad$ Jiayu Wang$^{1\ast}$ \quad
  Taylor Berg-Kirkpatrick$^2\quad$}\\
\makebox[\textwidth][c]{\textbf{Nadia Polikarpova}$^2\quad$ \textbf{Loris D'Antoni}$^1\quad$}\\
  $^1$University of Wisconsin-Madison \quad\quad
  $^2$University of California San Diego\\
 {\texttt{\{kpark247, jwang2782, ldantoni\}@wisc.edu}, ~ 
  \texttt{\{tberg, npolikarpova\}@ucsd.edu}}
}

\input{defs}

\begin{document} 
\maketitle

\begin{abstract}
Large Language Models (LLMs) struggle with reliably generating highly structured outputs, such as program code, mathematical formulas, or well-formed markup.
Constrained decoding approaches mitigate this problem by greedily restricting what tokens an LLM can output at each step to guarantee that the output matches a given constraint.
Specifically, in \emph{grammar-constrained decoding} (GCD),
the LLM's output must follow a given grammar.
In this paper we demonstrate that GCD techniques (and in general constrained decoding techniques) can \emph{distort the LLM's distribution}, 
leading to outputs that are grammatical but appear with likelihoods that are not proportional to the ones given by the LLM, and so ultimately are low-quality.
We call the problem of aligning sampling with a grammar constraint, \emph{grammar-aligned decoding} (GAD),
and propose \emph{adaptive sampling with approximate expected futures} (ASAp), a decoding algorithm that guarantees the output to be grammatical while provably producing outputs that match the conditional probability of the LLM's distribution conditioned on the given grammar constraint.
Our algorithm uses prior sample outputs to soundly overapproximate the future grammaticality of different output prefixes.
Our evaluation on code generation and structured NLP tasks shows how ASAp often produces outputs with higher likelihood (according to the LLM's distribution)
than existing GCD techniques, while still enforcing the desired grammatical constraints. 
\footnote{Our code, datasets, and checkpoints are available at: \url{https://github.com/ebmoon/transformers-GAD}.}
\end{abstract}

\input{intro}

\input{problem}

\input{algo}
\input{eval}
\input{related}

\subsection*{Acknowledgement}
The authors would like to thank NeurIPS anonymous reviewers for their insightful feedback and helpful discussions. The authors thank Gurindar S. Sohi, Shivaram Venkataraman, Ming Liu, and Yixuan Li for the support of computing resources.
This work is supported, in part, by
NSF under grants CCF-1918211, CCF-1955457, CCF-2023222, CCF-2200333, CCF-2211968, CCF-2402833, CCF-2422214, and CCF-2446711.
Any opinions, findings, and conclusions or recommendations expressed in this publication are those of the authors, and do not necessarily reflect the views of the sponsoring entities.
The authors thank Pawe\l{} Parys for finding and fixing errors in the proof of Theorem 2.

\bibliography{main}

\newpage
\appendix
\begin{center}
	\textbf{\LARGE Appendix }
\end{center}
\input{appendix}

\newpage\  
\newpage

\end{document}

%% file: defs.tex
\definecolor{npcolor}{RGB}{255,0,255}
\definecolor{tbcolor}{RGB}{0,150,0}
\definecolor{ldcolor}{RGB}{200,30,50}

\newtheorem{example}{Example}

\newtheorem{theorem}{Theorem}

\newcommand{\ie}{\emph{i.e.\@}\xspace}

\newcommand{\eg}{\emph{e.g.\@}\xspace}

\newcommand{\wrt}{\emph{wrt.\@}\xspace}

\newcommand{\sygus}{\textsc{SyGuS}\xspace}

\newcommand{\samples}{S}
\newcommand{\indicator}{\mathbbm{1}}
\newcommand{\grammar}{\mathcal{G}}
\newcommand{\nonterms}{\mathcal{N}}
\newcommand{\terms}{\Sigma}
\newcommand{\lang}{\mathcal{L}}
\newcommand{\langpre}{\lang_{\mathrm{prefix}}}
\newcommand{\ruleset}{\mathcal{R}}
\newcommand{\start}{S}
\newcommand{\nterm}{A}
\newcommand{\sent}{w}
\newcommand{\ev}{\mathbb{E}}
\newcommand{\prob}{P}
\newcommand{\probgrammar}{Q}
\newcommand{\probgrammarpg}[2]{\probgrammar^{{#1},{#2}}}
\newcommand{\probgcd}{\tilde\probgrammar_{\mathrm{GCD}}}
\newcommand{\evgrammar}{c}
\newcommand{\tevgrammar}{\tilde c}
\newcommand{\step}{\Rightarrow}
\newcommand{\manystep}{\Rightarrow^*}

\lstdefinelanguage{nolang}{
	literate=%
    {->}{$\rightarrow$}2
    {=.=}{$\doteq$}1
    {==}{$=$}1
    {!=}{$\neq$}1
    {&&}{$\land$}1
    {||}{$\lor$}1
    {<}{$<$}1
    {>}{$>$}1
    {<=}{$\le$}1
    {>=}{$\ge$}1
	,
	numbers=none,
	basicstyle=\ttfamily,
	commentstyle=\itshape\color{commentgreen},
	keywordstyle=\bfseries,
	ndkeywordstyle=\bfseries
}

\newcommand{\T}[1]{\mbox{\lstinline[language=nolang,basicstyle=\ttfamily,columns=fixed]^#1^}}

%% file: intro.tex
\section{Introduction}\label{sec:intro}

Despite their remarkable success,
pre-trained Large Language Models (LLMs) often struggle with generating highly structured outputs, 
such as program code, configuration files, or mathematical formulas.
A na\"ive approach to enforcing structure is \emph{rejection sampling},
which repeatedly samples strings from the LLM and checks them against a validity oracle,
typically in the form of a \emph{context-free grammar} (CFG).
Rejection sampling is highly inefficient or simply intractable
for restrictive grammars and long output sequences---i.e., most generated strings will not be in the target grammar.

Constrained decoding addresses the inefficiency of rejection sampling by greedily  ``forcing'' the LLM output to satisfy the given constraint. 
Specifically, when the constraint is given as a grammar, \emph{grammar-constrained decoding} (GCD)~\cite{geng2024grammarconstrained,wang2023grammar,willard2023efficient},
can build automata that allow for on-the-fly masking of tokens that will provably lead to outputs outside of the grammar during decoding.

While GCD does not incur the overhead of rejection sampling---i.e., the generated output is always in the language of the grammar---we show that GCD and in general all forms of structured decoding introduce a new problem:
\textbf{structured decoding distorts the LLM's learned language distribution},
effectively hindering the LLM's capabilities.

This paper introduces and formalizes \emph{grammar-aligned decoding} (GAD),
the problem of sampling from an LLM
so that the outputs 
\begin{enumerate*}[label=(\arabic*)]
    \item are guaranteed to adhere to a given grammar, and
    \item are unbiased \wrt the LLM's distribution.
\end{enumerate*}
Although exact GAD is intractable in general (similar to rejection sampling),
we propose a new adaptive decoding algorithm for \emph{approximate} GAD,
which starts off as GCD and gradually converges to the LLM's distribution,
and thus allows trading off between efficiency and accuracy.
The algorithm, which we dub \textit{Adaptive Sampling with Approximate Expected Futures} (ASAp),
is built ``on top'' of existing constrained decoding algorithms.
Whereas GCD approaches simply mask out tokens that lead to non-grammatical sequences for a given prefix, ASAp remembers  for all sampled prefixes the probability associated with masked-out tokens and uses it to upper bound the probability of grammaticality.
By updating this bound when more samples are observed, the decoding algorithm converges to the desired probability distribution---i.e., it samples outputs from the LLM-induced probability conditioned on the outputs being accepted by the grammar.
The idea works for any structured decoding approach and not just for GCD, but in this paper we focus our evaluation on constraints expressed via grammars.

We evaluate ASAp on two structured prediction tasks:
formal program synthesis and constituency parsing.
Our experiments on program synthesis and NLP tasks show 
that GCD techniques generate outputs that are grammatical but unlikely according to the LLM, while with ASAp, the likelihood of the generated outputs improves over time, converging to the target constrained LLM---i.e., GAD better respects the LLM while still enforcing the constraints.

%% file: problem.tex
\section{Grammar-Aligned Decoding}\label{sec:problem}

In this section, we formalize the problem of \emph{grammar-aligned decoding} (GAD)
as decoding from an autoregressive language model while enforcing the output sequence to be accepted by a given context-free grammar.
We also demonstrate the limitations of existing approaches to this problem.

\paragraph{Language Models}

An (autoregressive) language model defines a probability distribution $\prob$ on the set of all strings $\sent \in \terms^*$ 
over a vocabulary of tokens $\terms$ 
via a product of left-to-right next-token conditional distributions $\prob(\sent_1 \ldots \sent_n) = \Pi^n_{i=1} \prob(\sent_i \mid \sent_{1:i-1})$.

\paragraph{Context-Free Grammars}

\begin{wrapfigure}{r}{0.41\textwidth}
\vspace{-10pt}
\[
\begin{array}{rl}
    S ::= & \T{00000} \mid \T{1} A_2   \\
    A_i ::= & \T{0} A_{i+1} \mid \T{1} A_{i+1}\textrm{, for}\ i=2,3,4 \\
    A_5 ::= & \T{0} \mid \T{1}
\end{array}
\]
\caption{CFG $\grammar_{sk}$ over tokens $\terms = \{0,1\}$,
written in Backus-Naur form (BNF) notation.
This grammar accepts the string \T{00000} and all length-5 strings that start with a 1.}
\label{fig:grammar}
\end{wrapfigure}

A \emph{context-free grammar} (CFG) is a quadruple $\grammar = (\terms, \nonterms, \start, \ruleset)$, 
where $\terms$ is a vocabulary of tokens (also called terminal symbols), 
$\nonterms$ is a finite set of non-terminal symbols,
$\start \in \nonterms$ is the starting non-terminal, 
and $\ruleset$ is the set of production rules.
An example CFG is shown in \autoref{fig:grammar}.
%
%
A grammar $\grammar$ defines a \emph{single-step derivation} relation on sequences of symbols $\alpha, \beta, \gamma \in (\nonterms \cup \terms)^*$:
$\alpha \nterm \gamma \step \alpha \beta \gamma$ if $\nterm \to \beta \in \ruleset$.
The reflexive transitive closure of this relation is called \emph{derivation} and written $\manystep$.
A sequence of tokens $\sent$ is a \emph{sentence} if it is derivable from $\start$;
the set of all sentences is called the \emph{language} of the grammar $\grammar$, that is,
$\lang(\grammar) = \{\sent \in \terms^* \mid \start \manystep \sent\}$.
The following example illustrates these definitions.

\vspace{1mm}
\begin{example}[CFG Derivations]
Given the CFG $\grammar_{sk}$ shown in \autoref{fig:grammar}, the string \T{00000} belongs to the language $\lang(\grammar_{sk})$ because it can be derived using the derivation $S \Rightarrow \T{00000}$.
The string \T{10101} is also in $\lang(\grammar_{sk})$ and can be derived as follows:
\[
S \Rightarrow \T{1}A_2 \Rightarrow \T{10}A_3 \Rightarrow \T{101}A_4 \Rightarrow \T{1010}A_5 \Rightarrow \T{10101}
\]
Each step replaces a nonterminal symbol using a production rule in $\grammar_{sk}$---e.g., in the string $\T{10}A_3$, the nonterminal $A_3$ is rewritten as $\T{1}A_4$ by applying the rule $A_3 \rightarrow \T{1}A_4$,
resulting in the string $\T{101}A_4$.
\end{example}

In addition, we define the \emph{prefix language} of $\grammar$ as the set of all prefixes of sentences in $\lang(\grammar)$:
$\langpre(\grammar) = \{\sent \in \terms^* \mid \sent v \in \lang(\grammar)\}$.


\paragraph{Grammar-Aligned Decoding}
Given a model distribution $\prob$ and a CFG $\grammar$,
\emph{grammar-aligned decoding (GAD)} is the task of sampling from the distribution $\probgrammarpg{\prob}{\grammar}$
that is \emph{proportional} to $\prob$ but \emph{restricted} to sentences in $\grammar$:
\[
\probgrammarpg{\prob}{\grammar}(\sent)  = \frac{\indicator[\sent \in \lang(\grammar)] \cdot \prob(\sent)}{\sum_{\sent'} \indicator[\sent' \in \lang(\grammar)] \cdot \prob(\sent')}
\]
When $\prob$ and $\grammar$ are clear from context, we will write $\probgrammar(\sent)$ instead of $\probgrammarpg{\prob}{\grammar}(\sent)$.

\begin{wrapfigure}{r}{0.5\textwidth}
\centering
\includegraphics[width=0.5\textwidth]{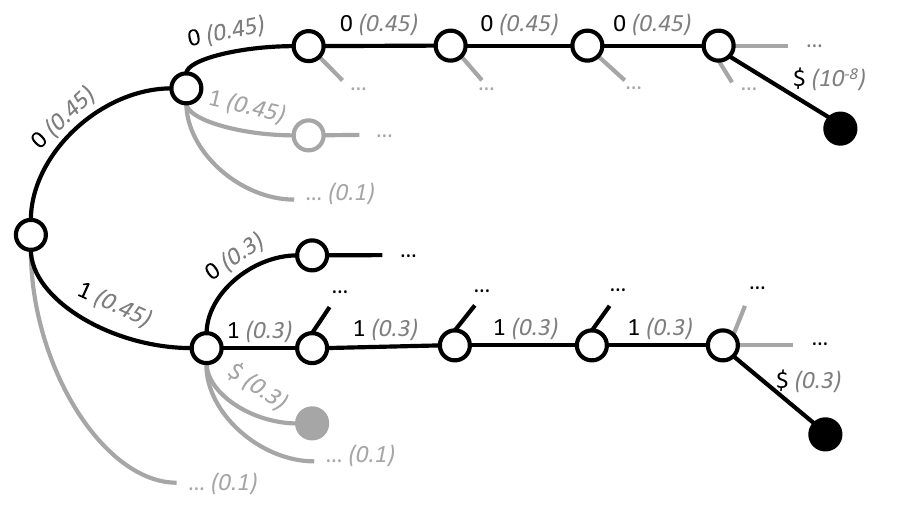}
\caption{Fragment of the conditional model distribution $\prob$ for \autoref{ex:problem} depicted as a trie.
Each node corresponds to a prefix $\sent_{1:i-1}$, and each edge is annotated with the next token $\sent_i$ and its conditional probability $\prob(\sent_i \mid \sent_{1:i-1})$.
Filled nodes are complete strings.
Grayed out parts of the trie are outside of the grammar $\grammar_{sk}$.
}\label{fig:model}
\end{wrapfigure}

\begin{example}[GAD]\label{ex:problem}
Consider the distribution $\prob$ that arises from prompting an LLM to ``generate a binary string that ends with a \T{1}''.
We expect $\prob$ to assign high probability to strings of the form $(\T{0} \mid \T{1})^* \T{1}$---%
\ie those that satisfy the prompt (\texttt{Mixtral-8x7B-Instruct-v0.1} (temperature=1) generates binary strings that end with a \T{1} around 90\% of the time.)
A snippet of a possible distribution $\prob$ is depicted in \autoref{fig:model}.
 
Suppose we constrain the model's output to the language of the grammar $\grammar_{sk}$ in \autoref{fig:grammar},
which only accepts strings of length 5.
Moreover, $\grammar_{sk}$ only accepts one string that starts with \T{0}, i.e., \T{00000}, which does not end with \T{1}.
In \autoref{fig:model}, the grayed out parts of the trie are tokens that lead to sequences outside of the grammar $\grammar_{sk}$.
According to the definition of GAD, 
the target sampling distribution $\probgrammarpg{\prob}{\grammar_{sk}}$ should assign:
\begin{enumerate*}[label=(\roman*)]
    \item high probability to all eight strings of the form $\T{1} \sent_2 \sent_3 \sent_4 \T{1}$---which conform both to the grammar and the prompt;
    \item low probability to the string \T{00000}---which conforms to the grammar but not the prompt; and
    \item zero probability to all other strings.
\end{enumerate*}
\end{example}

\paragraph{Exact GAD}
Can one exactly sample from $\probgrammarpg{\prob}{\grammar}$? 
Rejection sampling, which repeatedly draws from $\prob$ until a sample lands in $\lang(\grammar)$, provably yields exact samples according to $\probgrammarpg{\prob}{\grammar}$, 
but if $\prob$ assigns most of its mass outside of $\lang(\grammar)$, it is intractably slow, especially if the prompt is not including information about the grammar (see~\cite{wang2023grammar}). 
For \autoref{ex:problem}, rejection sampling would be highly inefficient
because the model would generate many strings that are not of length five.

In contrast, exact sampling from $\prob$ is efficient because its joint distribution is represented by a product of easily computed left-to-right conditionals, 
enabling ancestral sampling (i.e., generating tokens left to right, conditioned on already generated tokens).
Can we similarly factor $\probgrammar$ into a product of left-to-right conditionals $\probgrammarpg{\prob}{\grammar}(\sent_i | \sent_{1:i-1})$,
to enable ancestral sampling?

For simplicity, let us assume that $\prob$ is a distribution over sequences of exactly length $n$
(although, in practice, language models can produce `stop' tokens which allow for a valid distribution on sequences of all lengths). 
The exact conditionals of $\probgrammarpg{\prob}{\grammar}$ are given by:
\begin{equation}
\label{eq:prob-gad-tokens}
\begin{array}{rcl}
\probgrammarpg{\prob}{\grammar}(\sent_i \mid \sent_{1:i-1}) & \propto & \sum_{\sent_{i+1:n}} \left[ \indicator[\sent \in \lang(\grammar)] \cdot \Pi^n_{j=i} \prob(\sent_j \mid  \sent_{1:j-1}) \right] \\
 & \propto & \prob(\sent_i \mid \sent_{1:i-1}) \cdot \ev_{\prob(\sent_{i+1:n} \mid \sent_{1:i})} [\indicator[\sent \in \lang(\grammar)]]
\end{array}
\end{equation}
Thus, exact left-to-right sampling from $\probgrammarpg{\prob}{\grammar}$ consists of sampling from model conditionals $\prob(\sent_i \mid \sent_{1:i-1})$, 
with an additional weighting term $\evgrammar(\sent_{1:i})= \ev_{\prob(\sent_{i+1:n} \mid \sent_{1:i})} [\indicator[\sent \in \lang(\grammar)]]$ that considers the grammar. 

We refer to $\evgrammar(\sent_{1:i})$ as \emph{expected future grammaticality} (EFG),
\ie the probability that a continuation of $\sent_{1:i}$ sampled from $\prob$ lands in $\lang(\grammar)$.
Using this notation, we can write the exact left-to-right sampling conditional explicitly as:
\begin{equation}
\label{eq:prob-gad-by-weight}
\probgrammarpg{\prob}{\grammar}(\sent_i \mid \sent_{1:i-1}) = \frac{\prob(\sent_i \mid \sent_{1:i-1}) \cdot c(\sent_{1:i})}{\sum_{\sent'_i} \prob(\sent'_i \mid \sent_{1:i-1}) \cdot \evgrammar(\sent_{1:i-1}, \sent'_i)}
\end{equation}

To see why computing this conditional is intractable, 
consider using dynamic programming to compute $\evgrammar(\sent_{1:i})$ by marginalizing over a product of potential functions: 
the set of model conditionals and an indicator potential for the grammar. 
While the indicator potential can be factorized across rules in the grammar,
the model's contribution generally does not factorize: 
in practice, the final conditional probability $\prob(\sent_n \mid \sent_{1:n-1})$ is a global potential function,
defined by a non-linear neural network touching every variable. 
Thus, the main goal of this paper is to develop effective approximations to the EFG $\evgrammar(\sent_{1:i})$, 
which would enable us to compute the left-to-right conditionals of $\probgrammar$.

\paragraph{Limitations of Grammar-Constrained Decoding}

Existing work~\cite{wang2023grammar, willard2023efficient} has proposed \emph{grammar-constrained decoding} (GCD) 
as a way to efficiently sample from an autoregressive language model subject to grammar constraints.
Although the exact details of these techniques vary depending on class of grammars they support,
the common thread is that they rely on an \emph{incremental parser},
which can efficiently check whether a given string $\sent$ is a prefix of a sentence in the grammar, i.e., 
$\sent \in \langpre(\grammar)$.
When given a sentence $\sent_{1:i-1}$, GCD techniques use this parser during decoding to mask out any next token $\sent_i$
that results in a prefix $\sent_{1:i}$ for which no completion will produce a sequence in the grammar.
%
Using the trie in \autoref{fig:model} as an example, one can think of GCD as sampling a path through the trie by selecting only among the black outgoing edges from every node,
proportional to their conditional probabilities in the diagram
(\eg the first token is \T{0} or \T{1} with equal probability).

In terms of the GAD problem,
we can view GCD as approximating the exact left-to-right conditionals $\probgrammarpg{\prob}{\grammar}(\sent_i \mid \sent_{1:i-1})$ by the conditional distribution $\probgcd(\sent_i \mid \sent_{1:i-1})$,
defined as follows:
\[
\probgcd(\sent_i \mid \sent_{1:i-1}) = \frac{\prob(\sent_i\mid \sent_{1:i-1}) \cdot \indicator[\sent_{1:i} \in \langpre(\grammar)]}{\sum_{\sent'_i} \prob(\sent'_i\mid \sent_{1:i-1}) \cdot \indicator[\sent_{1:i-1},\sent'_i \in \langpre(\grammar)]}
\]
Though not originally formulated in this way, we can view recent work on GCD~\cite{wang2023grammar, willard2023efficient} as forming a binary approximation $\indicator[\sent_{1:i} \in \lang_{\textrm{prefix}}(\grammar)]$ 
to the EFG $\evgrammar(\sent_{1:i})$.
In other words, while GCD considers the \textit{possibility} of future grammaticality, 
it makes no attempt to integrate the model's likelihood to estimate \textit{expected} future grammaticality,
which can lead to substantial bias in the sampling distribution---i.e., every EFG such that $\evgrammar(\sent_{1:i})>0$ will simply be approximated via the value 1.


\begin{example}[GCD]\label{ex:gcd-00000}
Consider again the GAD problem from \autoref{ex:problem},
where our target sampling distribution $\probgrammarpg{\prob}{\grammar_{sk}}$ assigns high probability to strings that both start and end with a \T{1}
and a low probability to the string \T{00000}.
However, we observe that GCD~\cite{Geng2023} generates strings ending with a \T{1} only 30\% of the time---%
i.e., GCD has effectively ruined the LLM's ability to follow the prompt by biasing sampling towards \T{00000}, an incorrect output.

When generating the first token (\T{0} or \T{1}), 
the GCD algorithm does not know how many grammatical strings can start with each character and, more importantly, \emph{how likely} these strings are under $\prob$. 
Since both tokens \T{0} and \T{1} have the possibility of leading to a grammatical string, 
GCD will estimate their expected future grammaticality as 1, and choose each of them roughly half of the time (since $\prob(\T{0}) \approx \prob(\T{1})$).
Once GCD has chosen \T{0}, however, it becomes ``trapped'' in the part of the search space
where the only grammatical string is the low-probability sequence \T{00000}.
\end{example}


\autoref{ex:gcd-00000} illustrates how existing GCD approaches can hinder the language model's abilities to explore the space of possible outputs according to the learned distribution, 
thus highlighting the importance of designing a better approximation to the EFG $\evgrammar(\sent_{1:i})$;
this is addressed in the next section.


%% file: algo.tex
\section{Adaptive Sampling with Approximate Expected Futures (ASAp)}\label{sec:algo}

In this section, we propose an adaptive sampling algorithm that iteratively builds better approximations of the future grammaticality of a sequence. Our procedure operates by sampling \textit{repeatedly}, each time bounding lost probability mass to provably ungrammatical areas of the search space in order to better guide the next sampling iteration. As a result, our algorithm converges over many iterations to exact samples from the constrained LLM distribution, allowing for a flexible trade-off between efficiency and accuracy. 



\begin{figure}[t]
    \begin{minipage}{0.5\textwidth}
        \includegraphics[width=1\textwidth]{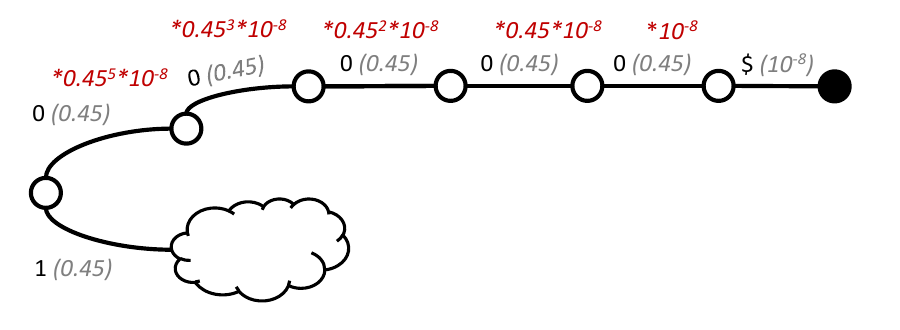}
    \end{minipage}%
    \begin{minipage}{0.5\textwidth}
        \includegraphics[width=1\textwidth]{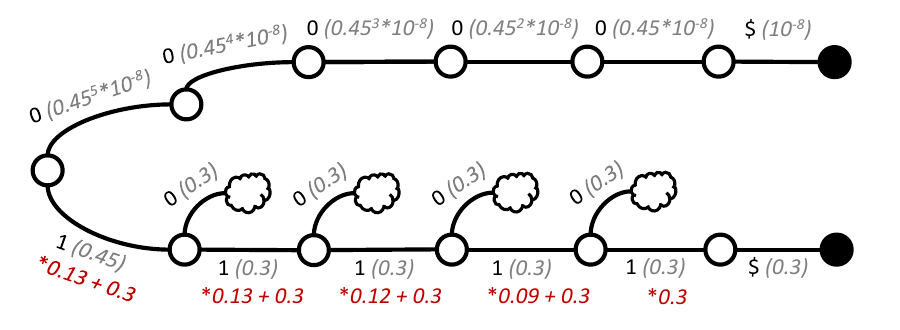}
    \end{minipage}
\caption{Illustration of the trie built by ASAp after sampling \T{00000} as the first string (left) and after sampling \T{11111} as the second string (right).
EFG updates after each iteration are shown in red.}\label{fig:samples}
\end{figure}

\paragraph{Overview of the Algorithm}
\label{se:left-to-right}
GCD approaches poorly approximate the desired distribution because they greedily sample prefixes without worrying about the EFG.
When sampling the first token in \autoref{ex:gcd-00000}, GCD simply uses the likelihood for tokens 0 and 1 assigned by the LLM without considering the probability that these next tokens would result in grammatical completions if sampling were unconstrained---i.e. without incorporating the critical EFG re-weighting terms that are necessary for unbiased sampling from the constrained LLM distribution.
However, if GCD ends up sampling \T{0} as the first token for \autoref{ex:gcd-00000}, it will necessarily sample the string \T{00000} since no other sequences starting with \T{0} are allowed by the grammar. We can ``learn'' from this result: the true probability mass assigned to all grammatical sequences starting with a \T{0} is not 0.45 as the LLM's next token probability would have us believe; instead, the total grammatical mass in this section of the search space is the joint probability of the single string \T{00000}, which is the much lower value of $0.45^5 * 10^{-8}$ as depicted in \autoref{fig:samples}. In other words, simply by sampling \T{00000}, we can better approximate (in this case, exactly) the EFG of tokens along this path.

The key insight behind our algorithm, which we call ASAp, is that we can iterate this process of discovering lost grammatical probability mass by repeatedly sampling and revising transition weights after each sample is produced. More formally, we can think of this procedure as starting with GCD's over-approximation to each EFG $\evgrammar(w_{1:i})$ term, and then, through repeated sampling and discovery of mass assigned to non-grammatical completions, reducing each overapproximation to make it more accurate. In the limit, the approximations converge to exact EFG estimates and unbiased sampling. 

Two possible first iterations of the ASAp algorithm are depicted in \autoref{fig:samples}.
%
In the first iteration (left of \autoref{fig:samples}), after sampling the sequence \T{00000}, the algorithm directly addresses the issue that arose in \autoref{ex:gcd-00000} by attempting to better approximate the probability mass of potential grammatical completions of each prefix of \T{00000} (red quantities).
For example, the expected future grammaticality of the prefix \T{0000} it is now $0.45*10^{-8}$---i.e., the algorithm effectively ``looks ahead'' to determine that only one valid (but low probability) string \T{0\$} that can follow \T{0000}.
%
The ideas developed in GCD allow us to efficiently compute, for a given string, the likelihood of the next tokens that will immediately result in non-grammaticality.

If we only sample one string from the LLM, we cannot hope to do better than GCD in terms of sampling faithfully in a grammar-aligned way.
However, if we were to now sample once more, we could now better direct our sampling strategy.
In the second iteration (right of \autoref{fig:samples}), the string \T{11111} is sampled and the expected future grammaticality is updated (red quantities).
Note that at this point the probabilites assigned to the string \T{00000} from the earlier iteration have already been updated.

By repeating the above approach multiple times (i.e., by producing more samples), the ASAp algorithm produces precise approximations of the expected future grammaticalities and thus better samples from the constrained LLM.

\paragraph{Algorithm Formalization}
\label{sec:algorithm-formalization}

The key quantity that the algorithm approximates based on past samples is the \textit{expected future grammaticality} (EFG)
$\evgrammar(w_{1:i})=\ev_{p(w_{i+1:n}|w_{1:i})} [\indicator[w \in \lang(\grammar)]]$. At iteration $m+1$, our algorithm uses the set of samples $\samples=\{s_1,\ldots, s_m\}$ observed so far to compute an overapproximation $\tevgrammar_\samples(w_{1:i})$ of $\evgrammar_\samples(w_{1:i})$ for every possible string $w_{1:i}$.
The overapproximation is inductively defined:
\begin{equation}
\label{eq:ev-grammar-recursive}
\begin{array}{ll}
\tevgrammar_\samples(w_{1:i}) = 
\indicator[w_{1:i} \in \langpre(\grammar)] & \text{no string in }\samples \text{ starts with } w_{1:i} \\
     \tevgrammar_\samples(w_{1:i}) = 
\sum_{w_{i+1}}
\prob(w_{i+1}\mid w_{1:i}) \cdot \tevgrammar_\samples(w_{1:i+1}) & \text{otherwise}
\end{array}
\end{equation}
Intuitively, if no samples in $\samples$ start with the prefix $w_{1:i}$, then $\tevgrammar_\samples(w_{1:i})$, the overapproximation of EFG is simply whether the string is or is not a valid prefix in the grammar---i.e. the same overapproximation used by GCD.
If, on the other hand, we \textit{have} encountered the prefix $w_{1:i}$ before in previous samples in $S$, the overapproximation uses the next token likelihoods that were computed during the previous sampling runs of the algorithm to compute a better estimate of EFG.

For example, in \autoref{fig:samples}, once we have sampled the sequences \T{00000} and \T{11111}, we have that $\tevgrammar_\samples(\T{0000})=0.45*10^{-8}$ and $\tevgrammar_\samples(\T{110})=1$ (i.e., we have not seen a sample with the prefix \T{110} yet).

\begin{theorem}
\label{thm:upper}
$\forall w_{1:i} \in \Sigma^\ast, \tevgrammar_\samples(w_{1:i})\geq\evgrammar(w_{1:i}) $.
\end{theorem}

\begin{proof}
  To see that $\tevgrammar_\samples(\sent_{1:i})$ is indeed an upper bound on $\evgrammar(\sent_{1:i})$, consider two cases: First, suppose $\sent_{1:i}$ is not a prefix of any string in $S$. In this case, $\evgrammar(\sent_{1:i}) \leq 1 =\tevgrammar_\samples(\sent_{1:i})$. Second, we need to prove that $\forall \sent_{1:i} \in \mathrm{prefix}(S), \evgrammar(\sent_{1:i})\leq\tevgrammar_\samples(\sent_{1:i})$, where $\mathrm{prefix}(S)$ is the set of (finitely many) prefixes of string in $S$. We proceed by induction, starting from the longest prefixes. No matter whether we are in the base case or in the inductive step, for any $\sent_{i+1}$, the string $\sent_{1:i+1}$ either is not a prefix of $S$, where by the first case we obtain $\evgrammar(\sent_{1:i+1})\leq\tevgrammar_\samples(\sent_{1:i+1})$, or it is a longer prefix of $S$, for which the same inequality holds by the induction hypothesis. Observing that $\evgrammar(\sent_{1:i})$ factorizes similarly to $\tevgrammar_\samples(\sent_{1:i})$, this leads us to the following inequality: 
  \begin{align*}
    \evgrammar(\sent_{1:i}) &= \sum_{\sent_{i+1}}\prob(\sent_{i+1}\mid \sent_{1:i}) \cdot\evgrammar(\sent_{1:i+1})\\
    &\leq \sum_{\sent_{i+1}}\prob(\sent_{i+1}\mid \sent_{1:i})\cdot\tevgrammar_\samples(\sent_{1:i+1})
    = \tevgrammar_\samples(\sent_{1:i})
  \tag*{\qedhere}\end{align*}
\end{proof}

The sampling procedure itself proceeds autoregressively like GCD, but using the iteratively updated EFG estimates we have just defined, $\tevgrammar_\samples$.
Specifically, the left-to-right sampling conditional for our procedure, $\tilde \probgrammar_\samples(w_i | w_{1:i-1})$, after having previously sampled the strings in $\samples$, is defined as follows:
\begin{equation}
\tilde \probgrammar_\samples(w_i | w_{1:i-1}) = \frac{\prob(w_i\mid w_{1:i-1}) \cdot \tevgrammar_\samples(w_{1:i})}{\sum_{w'_i} \prob(w'_i\mid w_{1:i-1}) \cdot \tevgrammar_\samples(w_{1:i-1},w'_i)}
\end{equation}

\begin{wrapfigure}{r}{0.55\textwidth}
    \vspace{-15pt} 
    \begin{minipage}{0.55\textwidth}
        \begin{algorithm}[H]
        \caption{ASAp algorithm}\label{alg:cap}
        \begin{algorithmic}
        \State Initialize $S := \{\}$, $\tevgrammar_\samples(\cdot) := 1$
        \For{$m \le M$} 
            \State Draw $w_{1:n} \sim \tilde \probgrammar_\samples$ via ancestral sampling
            \State $S := S \cup \{ w_{1:n} \}$
            \For{$i$ in $(n-1) \ldots 1$}
                \For{$w'$ in $\{w' \mid w_{1:i} \cdot w' \notin \langpre(\grammar)\}$}
                    \State $\tevgrammar_\samples(w_{1:i} \cdot w') := 0$
                \EndFor
                \State $\tevgrammar_\samples(w_{1:i}) \hspace{-1px}:=\hspace{-1px} \sum_{w'} \hspace{-1.5px} \prob(w'\hspace{-1px}\mid\hspace{-1px} w_{1:i}) \cdot \tevgrammar_\samples(w_{1:i} \hspace{-1px}\cdot\hspace{-1px} w')$
            \EndFor
        \EndFor\\
        \Return Final sample $w_{1:n}$
        \end{algorithmic}
        \end{algorithm}
    \end{minipage}
\end{wrapfigure}
Our overall algorithm, which is presented in Algorithm~\ref{alg:cap}, then proceeds iteratively, using past samples to improve subsequent samples.
Whenever the sample set $S$ is updated with a new sample $w_{1:n}$, the overapproximation $\tevgrammar$ is updated for the prefixes of $w_{1:n}$. The update begins at the end of the sequence and proceeds backward toward the start, by the recursive definition in \autoref{eq:ev-grammar-recursive}.
In the listing, we assume that we are only interested in the final sample, but in our evaluation we will analyze whether the algorithm induces the desired distribution.

Next we provide a proof that this algorithm converges to exact estimates of EFG in the limit of infinite iterations, and therefore to exact samples from the constrained LLM distribution.
The theorem assumes almost sure termination of ancestral sampling in the unconstrained LLM distribution $P$---i.e., the LLM eventually terminates. 
\begin{theorem}\label{thm:convergence}
    Let $\samples_m = \{s_1,\ldots,s_m\}$ be the set of recorded samples up to the $m$-th iteration of ASAp. Then, $\forall \sent_{1:i}\in\langpre, \tevgrammar_{\samples_m}(\sent_{1:i}) \to \evgrammar(\sent_{1:i})$ as $m \to \infty$ almost surely.
\end{theorem}
\begin{proof}
Fix an arbitrary sequence $\sent_{1:i}\in\langpre$. Let $\lang(\sent_{1:i})$ be the set of all strings $\sent$ in $\lang$ extending the fixed prefix $\sent_{1:i}$. For every $m$, let $X_m\subseteq\langpre$ be the set of shortest extensions $\sent_{1:j}\in\langpre$ of $\sent_{1:i}$ that \textit{have not yet been encountered} in the first $m$ samples, that is, are not prefixes of any string in $\samples_m$. It can be easily seen that
\begin{align}
    \tevgrammar_{\samples_m}(\sent_{1:i})-\evgrammar(\sent_{1:i})&=\sum_{\sent_{1:j}\in X_m} \prob(\sent_{i+1:j}\mid \sent_{1:i}) \cdot\big(1-\evgrammar(\sent_{1:j})\big)\nonumber\\
    &\leq\sum_{\sent_{1:j}\in X_m}\prob(\sent_{i+1:j}\mid \sent_{1:i}),\label{eq:bound}
\end{align}
Now consider an arbitrarily small $\varepsilon>0$. Because the infinite sum $\sum_{\sent\in\lang(\sent_{1:i})}\prob(\sent_{i+1:\infty}\mid \sent_{1:i})$ equals $1$, there exists a \emph{finite} subset $V_\varepsilon\subseteq\lang(\sent_{1:i})$ such that $\sum_{\sent\in V_\varepsilon}\prob(\sent_{i+1:\infty}\mid \sent_{1:i})>1-\varepsilon$ (we just take appropriately long initial prefix of the infinite sum); in other words, $\sum_{\sent\in\lang(\sent_{1:i})\setminus V_\varepsilon}\prob(\sent_{i+1:\infty}\mid \sent_{1:i})<\varepsilon$.
Suppose that for some $m$ we have the property that for every $\sent\in V_\varepsilon$,
\begin{enumerate}
    \item if $\sent\in\lang$, then $\sent\in\samples_m$, and
    \item if $\sent\not\in\lang$, then its longest prefix $\sent_{1:i}$ that belongs to $\langpre$ is a prefix of some $\sent'\in\samples_m$.
\end{enumerate}
Coming back to \autoref{eq:bound}, we observe that the sum $\sum_{\sent_{1:j}\in X_m}\prob(\sent_{i+1:j}\mid \sent_{1:i})$ can be rewritten as the sum of $\prob(\sent_{i+1:\infty}\mid \sent_{1:i})$ over all strings $\sent\in\lang(\sent_{1:i})$ having a prefix in $X_m$. But strings $\sent\in V_\varepsilon\cap\lang$ do not have a prefix in $X_m$, because by (1) they were already sampled. Strings $\sent\in V_\varepsilon\setminus\lang$ also do not have a prefix in $X_m$, because by (2) all their prefixes were either encountered or are not in $\langpre$ (while $X_m\subseteq\langpre$). Thus, by the definition of $V_\varepsilon$, the sum $\sum_{\sent_{1:j}\in X_m}\prob(\sent_{i+1:j}\mid \sent_{1:i})$ is bounded by $\varepsilon$. 
Having this for every $\varepsilon>0$, we obtain convergence.

It remains to prove that indeed almost surely there exists a value of $m$ such that (1) and (2) hold for every word in $V_\varepsilon$. Because $V_\varepsilon$ contains only finitely many strings, it is enough to prove this for every string $\sent\in V_\varepsilon$ (and then take a maximum of the finitely many obtained values of $m$). Thus, fix such a string $\sent\in V_\varepsilon$; also fix a step number $m$, together with the current set of previous samples $\samples_m$. In case (1) (i.e., when $\sent\in\lang$), let $n=\mathit{len}(w)$, and in case (2) (i.e., when $\sent\not\in\lang$), let $n$ be the length of the longest prefix $\sent_{1:n}$ of $\sent$ that belongs to $\langpre$. In both cases, we need to prove that the probability that the string sampled in the next step will start with $\sent_{1:n}$ is not smaller than some positive constant, independent of $m$ and $S_m$ (for case (1) notice that after sampling the first $\mathit{len}(w)$ letters of $w$ we have no choice, only $w$ can be sampled). Take thus some $j\in\{1,\dots,n\}$, suppose that we have already sampled $w_{1:j-1}$, and we are now going to sample the $j$-th letter of the word. Once again, because there are finitely many positions $j$ in consideration, it is enough to prove that for each such $j$ the probability of sampling $w_j$ as the next letter is not smaller than some positive constant. But notice that $\tevgrammar_{\samples_m}(w_{1:j})\geq\evgrammar(w_{1:j})>0$ (the second inequality because $w_{1:n}\in\langpre$) and $\tevgrammar_{\samples_m}(w_{1:j-1}\cdot a)\leq 1$ for every letter $a$. It follows that the probability of sampling $w_j$ as the $j$-th letter, that is $\tilde \probgrammar_\samples(\sent_j | \sent_{1:j-1})$, is at least
\[\frac{\prob(\sent_i\mid \sent_{1:i-1}) \cdot\evgrammar(\sent_{1:i})} {\sum_{\sent'_i}\prob(\sent'_i\mid \sent_{1:i-1})}\,,\]
which is a positive constant independent from $m$ and $\samples_m$, as needed.
\end{proof}

%% file: eval.tex
\section{Experiments}\label{sec:eval}

We implemented the ASAp algorithm as an extension of the Transformers-CFG implementation of GCD~\cite{Geng2023}.
When the LLM generates a sequence $w_{1:n}$, the ASAp algorithm keeps track of the original LLM's probability $\prob(w_i \mid w_{1:i-1})$ for $1 \leq i \leq n$ and the set of allowed next tokens $\{w_i \mid w_{1:i-1}, w'_i \in \lang(\grammar) \}$ determined by the incremental parser in the Transformers-CFG library.
After the LLM finishes generating a sequence, our implementation of ASAp updates the overapproximation $\tevgrammar_\samples$ from the end of sequence by back-propagating the quantity 1 minus probability of the tokens that will for sure lead to non-grammatical sequences. 
The implementation of ASAp updates $\tevgrammar_\samples(w_{1:n-1}, w'_n)$ for all possible tokens $w'_n$, and then moves on to update $\tevgrammar_\samples(w_{1:n-2}, w'_{n-1})$ $\ldots$, $\tevgrammar_\samples(w_1, w'_2)$, $\tevgrammar_\samples(w'_1)$ using Equation \eqref{eq:ev-grammar-recursive}.

\begin{figure}
    \centering
    \begin{subfigure}{0.45\textwidth}
        \centering
        \input{codes/sygus_slia}
        \vspace{-0.5em}
        \caption{\T{SLIA/initials-small}}
        \label{fig:sygus-slia}
    \end{subfigure}
    \hspace{1.5em}\vrule\hspace{1em}
    \begin{subfigure}{0.45\textwidth}
        \centering
        \input{codes/sygus_bv4}
        \vspace{-0.5em}
        \caption{\T{INV-BV/find_inv_bvsle_bvlshr1_4bit}}
        \label{fig:sygus-bv4}
    \end{subfigure}

    \begin{subfigure}{0.45\textwidth}
        \centering
        \input{codes/grammar_slia}
        \vspace{-1em}
        \caption{Grammar for \T{f}}
        \label{fig:grammar-slia}
    \end{subfigure}
    \hspace{1.5em}\vrule\hspace{1em}
    \begin{subfigure}{0.45\textwidth}
        \centering
        \input{codes/grammar_bv4}
        \vspace{-1em}
        \caption{Grammar for \T{inv}}
        \label{fig:grammar-bv4}
    \end{subfigure}
       
    \caption{(a) A SLIA problem in which the grammar for the target function is explicitly defined.
    (b) INV-BV problem in which the grammar for the target function \T{inv} is implicitly defined.
    (c) The explicitly defined grammar for \T{f} written in BNF notation.
    (d) The implicitly defined grammar for \T{inv} written in BNF notation. The grammar is implicitly defined by primitives of BV logic and parameters of \T{inv}.
    The goal of each problem is to find an implementation for \textbf{\T{synth-fun}} functions that satisfies all the \textbf{\T{constraint}}s within a specified grammar---i.e., to find implementation of \T{f} in the grammar (c) and \T{inv} in the grammar (d).}
    \label{fig:sygus-example}
\end{figure}

\paragraph{Datasets and Models.}
We consider the benchmark from  Example~\ref{ex:gcd-00000} and three structured-decoding tasks.
Two of our tasks involve solving Syntax-Guided Synthesis Problems (SyGuS)~\cite{alur2019syguscomp}.
SyGuS is a standardized format where one provides a logical specification and a context free grammar of first-order terms and the goal is to synthesize a term in the grammar that satisfies the specification.
SyGuS is a natural fit for GAD and we consider two tasks from the standard SyGuS benchmarks where grammars vary from benchmark to benchmark: strings with linear integer arithmetic (SLIA) and conditional inverse generation with bit-vector arithmetic (INV-BV).
In the former, the grammar is used to restrict what constant strings one can use when building string-manipulating programs and in the latter the grammar is used to restrict constant bit-vectors and operations used to build invariants.
\autoref{fig:sygus-example} provides examples of SLIA and INV-BV problems.
For both families of benchmarks, our prompts consist of 3 in-context examples of the form (specification, solution) and the grammar is then provided as a constraint for GAD.
Our third task is the constituency parsing (CP) task already used in prior GCD work~\cite{geng2024grammarconstrained} where the grammar is used to help the model produce well-parenthesized parse trees for English sentences.

Due to constrained resources and needing to run inference multiple times to measure whether the distribution $\tilde \probgrammar$ is faithful to $\probgrammar$, we randomly select 15 SLIA problems, 15 INV-BV problems, and 6 CP problems.
We select the open-source Mistral-7B~\cite{jiang2023mistral} for evaluation due to its superior reasoning and code generation capabilities.

\paragraph{Measures.}
We run both algorithms for 2,000 iterations/sample on each benchmark.

To assess converge to the target distribution, we measure the Kullback–Leibler (KL) divergence between the distributions of GCD and ASAp  from the target distribution $\probgrammar$ for a given number of samples.
Because the ideal GAD distribution $\probgrammar_{\prob, \grammar}$ is proportional to the original LLM's distribution $\prob$ for sequences allowed by a grammar $\grammar$, we can use the LLM's distribution $\prob$ on all observed samples as an estimate $\probgrammar_{\prob, \grammar}$.
The quantity $KL(\probgrammar \| \prob)$ only differs by a constant from the KL divergence between empirical distributions and the ideal GAD distribution:
\[
KL(\tilde\probgrammar \| \prob) {=}\ev_{\tilde\probgrammar}\left[ \log \frac{\tilde\probgrammar}{\prob} \right]
{=} \ev_{\tilde\probgrammar}\left[ \log \frac{\tilde\probgrammar}{C {\cdot} \probgrammar_{\prob, \grammar}} \right]
{=} \ev_{\tilde\probgrammar}\left[ \log \frac{\tilde\probgrammar}{\probgrammar_{\prob, \grammar}} \right] - \log C {=}  KL(\tilde\probgrammar \| \probgrammar_{\prob, \grammar}) - \log C
\]
where $C = \sum_{w} \indicator[w \in \lang(\grammar)] \prob(w)$.
Thus, $KL(\tilde\probgrammar \| \prob)$ can be used to quantify the alignment between the empirical distributions of GCD and ASAp with the ideal GAD distribution.
 
\begin{figure}[t]
\begin{subfigure}[valign=t]{0.33\textwidth}
\includegraphics[width=\textwidth]{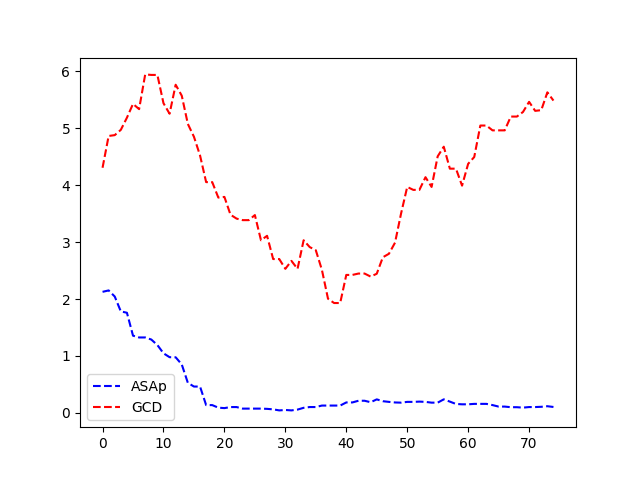}
\caption{Illustrative \autoref{ex:gcd-00000}}
\label{fig:binary-kl}
\end{subfigure}
\begin{subfigure}[valign=t]{0.33\textwidth}
\includegraphics[width=\textwidth]{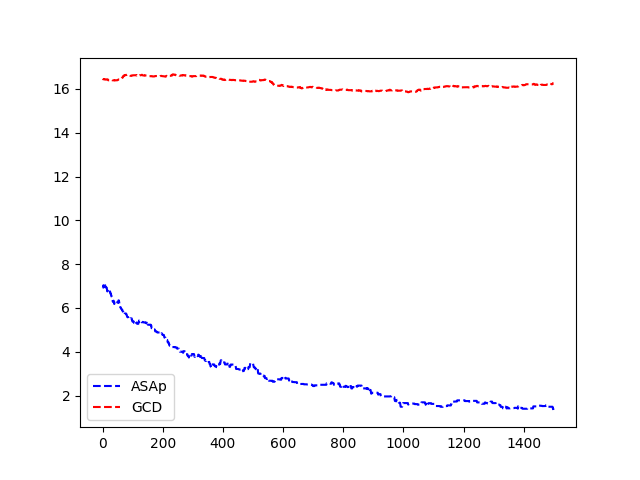}
\caption{\texttt{SLIA/name-combine-4-long}}
\label{fig:good-kl}
\end{subfigure}
\begin{subfigure}[valign=t]{0.33\textwidth}
\includegraphics[width=\textwidth]{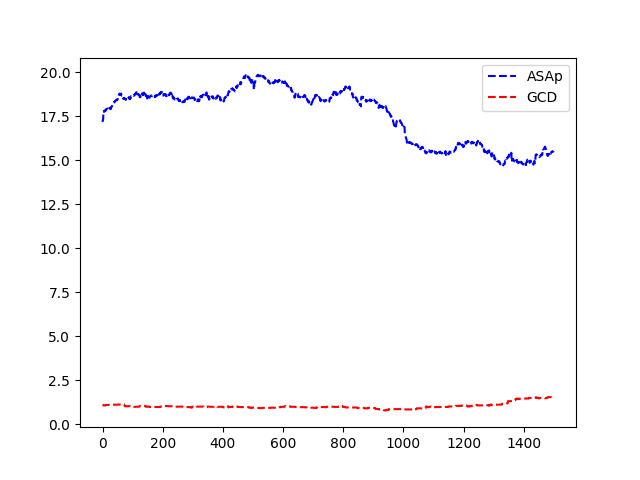}
\caption{\texttt{SLIA/initials-small}}
\label{fig:bad-kl}
\end{subfigure}
\caption{$KL(\probgrammar_{ASAp} \| \prob)$ and $KL(\probgrammar_{GCD} \| \prob)$}
\label{fig:kl-results}
\end{figure}

For example, \autoref{fig:binary-kl} shows convergence results for the first 75 iterations on the illustrative \autoref{ex:gcd-00000}---i.e., the KL divergence for $\tilde{\probgrammar}_{ASAp}$ quickly converges to 0 whereas the one for $\tilde{\probgrammar}_{GCD}$ doesn't.

\begin{figure}[t]
\begin{subfigure}[valign=t]{0.33\linewidth}
\includegraphics[width=\textwidth]{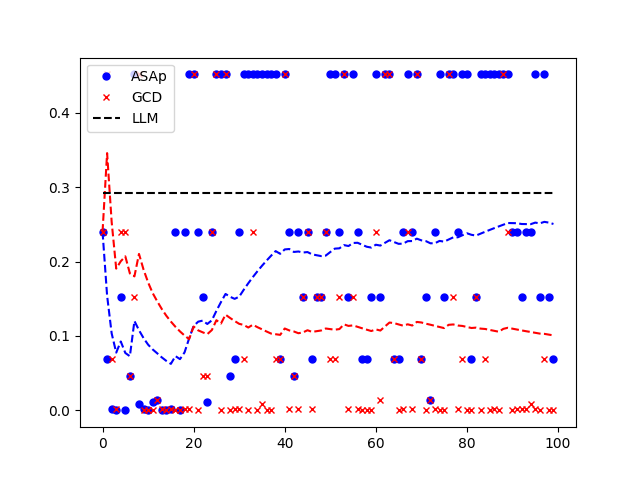}
\caption{Illustrative \autoref{ex:gcd-00000}}
\label{fig:binary-prob}
\end{subfigure}
\begin{subfigure}[valign=t]{0.33\linewidth}
\includegraphics[width=\textwidth]{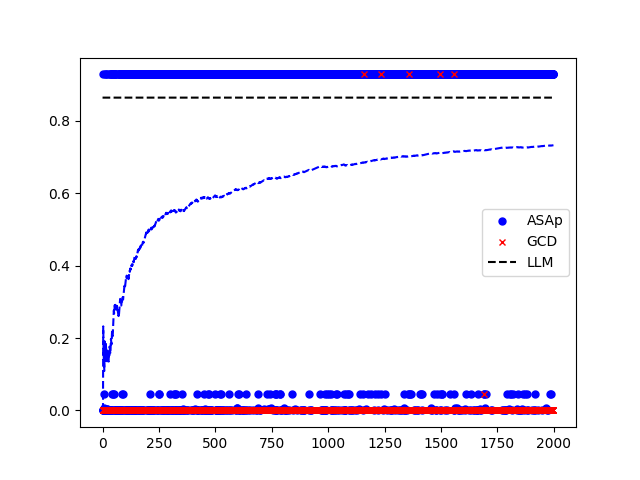}
\caption{\texttt{SLIA/name-combine-4-long}}
\label{fig:good-prob}
\end{subfigure}
\begin{subfigure}[valign=t]{0.33\linewidth}
\includegraphics[width=\textwidth]{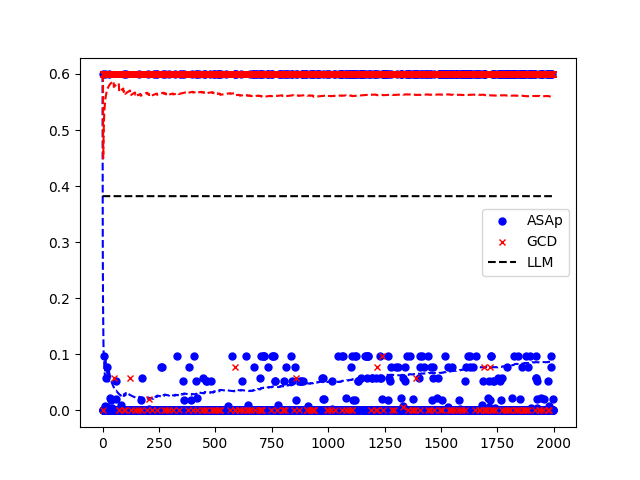}
\caption{\texttt{SLIA/initials-small}}
\label{fig:bad-prob}
\end{subfigure}
\caption{Expectations of $\tilde{\probgrammar}_{ASAp}$, $\tilde{\probgrammar}_{GCD}$, and $\prob$}
\label{fig:prob-results}
\end{figure}

We also compare the empirical expectations of the variables $\tilde{\probgrammar}_{GCD}$, $\tilde{\probgrammar}_{ASAp}$, and $\prob$.
For example, \autoref{fig:binary-prob} shows convergence results for the first 75 iterations on the illustrative \autoref{ex:gcd-00000}---i.e., $\tilde{\probgrammar}_{ASAp}$ converges to the right expectation.


\paragraph{Results.}
\autoref{fig:good-kl} and \autoref{fig:good-prob} illustrate a benchmark in which our ASAp algorithm quickly converges to the target distribution.
\autoref{fig:kl-results} depicts the KL divergence of a sliding window of size 500 (e.g., the points at x=800 denote the KL divergence of the samples 800-1300).
\autoref{fig:prob-results} depicts all the samples from the experiment, as well as how the expectations converges (a point at x=$i$ denotes the empirical expectation on the first $i$ samples.
For this case the expecation for GCD stays very close to 0.
 
Similarly, \autoref{fig:bad-kl} and \autoref{fig:bad-prob} illustrate a benchmark in which our ASAp algorithm converges slowly. 
In this case, bot ASAp and GCD are far from the target expectation (\autoref{fig:bad-prob}), but because GCD happens to be biased towards the most likely outcome, it exhibits better KL divergence.
The complete set of plots is shown in \autoref{app:plots}.

\begin{figure}[t]
\begin{subfigure}[valign=t]{0.33\textwidth}
\includegraphics[width=\textwidth]{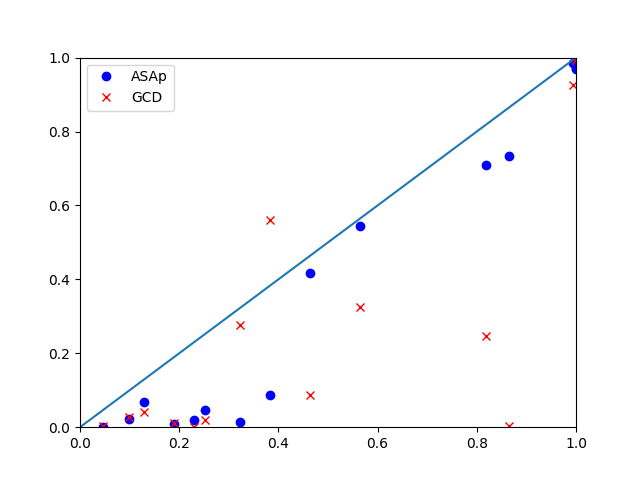}
\caption{SLIA}
\label{fig:scatter-slia}
\end{subfigure}
\begin{subfigure}[valign=t]{0.33\linewidth}
\includegraphics[width=\textwidth]{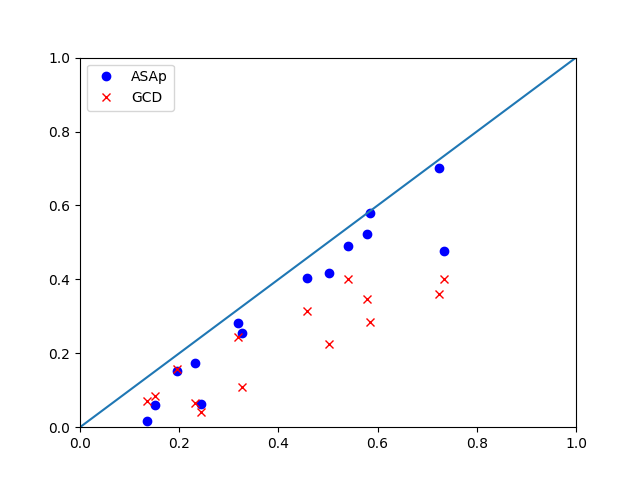}
\caption{INV-BV4}
\label{fig:scatter-bv4}
\end{subfigure}
\begin{subfigure}[valign=t]{0.33\linewidth}
\includegraphics[width=\textwidth]{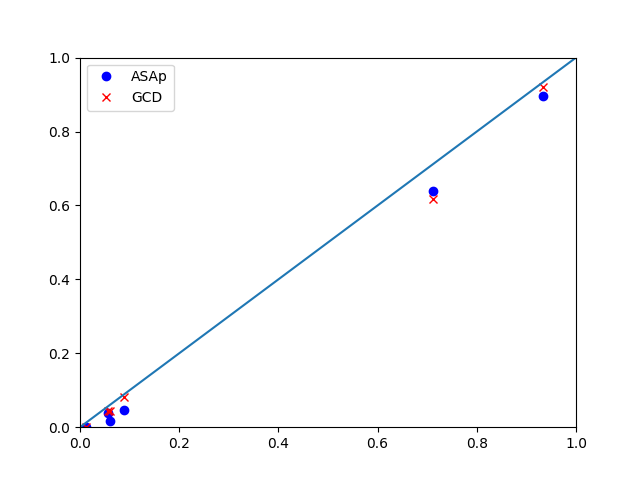}
\caption{CP}
\label{fig:scatter-cp}
\end{subfigure}
\caption{Scatter plots  of $\tilde{\probgrammar}_{ASAp}$ (\textcolor{blue}{$\bullet$}) and $\tilde{\probgrammar}_{GCD}$ (\textcolor{red}{$\times$}) vs. expectations of $\prob$ after 2,000 samples. Proximity to the diagonal indicates proximity to the actual expectation---e.g., a \textcolor{blue}{$\bullet$} at (0.45,0.4) indicates a benchmark where the empirical expectation of $\prob$ was 0.45 and $\tilde{\probgrammar}_{ASAp}$ had converged to an expectation of 0.4 after 2,000 iterations.}
\label{fig:scatter-plots}
\end{figure}

To better understand how the algorithms respectively converge, \autoref{fig:scatter-plots} plot for each benchmark category the expectations for each benchmark computed by GCD and ASAp against the target expectation of $\prob$ after 2,000 iterations.
The sum of least square difference between expectations computed by GCD and the expectations of $\prob$ are $2.259$ (SLIA), $1.852$ (INV-BV4), and $0.109$ (CP).
The sum of least square difference between expectations computed by ASAp and the expectation and those of $\prob$ are $1.242$ (SLIA), $0.802$ (INV-BV4), and $0.159$ (CP).
While we have too few points for CP to draw conclusions, the expectations computed by ASAp are much closer to the ones computed by GCD across our experiments.

While our work is interested in the theoretical convergence of the ASAp algorithm, we also report how the GCD and ASAp differ for solving the SLIA and INV-BV4 tasks---i.e., how many of the sampled programs are correct solutions to the given problem.
GCD and ASAp solve approximately the same set of problems (there is just one SLIA benchmark for which ASAp returns a valid solution on one sample and GCD never does so). 
ASAp produces correct samples 38\% more often than GCD (geomean), whereas for SLIA benchmarks that both tools can solve,  ASAp produces correct samples 73\% less often than GCD (geomean). 
Detailed results can be found in \autoref{app:correctness-sygus}.
These results are in line with the fact ASAp shows faster convergence on INV-BV4 benchmarks. 
For example, for the benchmark illustrated in \autoref{fig:good-kl}, ASAp returns the correct solution for 1588 samples, whereas GCD only returns the correct solution 12 times, whereas for the benchmark in \autoref{fig:bad-kl}, ASAp returns the correct solution 69 times and GCD 363 times.

\paragraph{Discussion and Limitations.}
As predicted by our theorems, on most benchmarks the ASAp algorithm converges to the desired distribution $\prob$ whereas GCD does not improve over time (i.e., it exhibits the bias described in this paper).

While ASAp has no strong effect on solving downstream tasks, we observe that on instances where the convergence is prominent, ASAp ends up sampling correct solutions more often than GCD, which is what we expect when the LLM has ``learned'' how to solve the given task.

The key limitation of our work is the current slow convergence of the ASAp algorithm. In some benchmarks, even after 2,000 iterations the KL divergence barely improves and even though the expectation of $\tilde{\probgrammar}_{ASAp}$ is improving, it converges very slowly.

We highlight that the contributions of this paper are discovering and formalizing the bias of existing constrained decoding approaches and proposing the first converging algorithm to address this problem. 
Now that we have identified the problem, there are many ``low-hanging fruits'' to improve our sampling strategy, which are great targets for future work---e.g., using forms of targeted beam search to bootstrap our sample set to better explore grammar paths and avoid sampling similar strings.

%% file: codes/sygus_slia.tex
\begin{lstlisting}[ tabsize=3, 
    basicstyle= \tt \footnotesize, 
    keywordstyle=\color{black}\bfseries, 
    commentstyle=\color{gray}, 
    escapeinside=``,
    language=Lisp,
    morekeywords = {set-logic, synth-fun, declare-var, constraint, check-synth, Int, Bool},
    deletekeywords = {replace},
    numbers = none
    ]
; Determines what terms can appear
(set-logic SLIA)
; The function to synthesize
(synth-fun f ((name String)) String
    ; The grammar for f to be synthesized in
    ((Start String (S))
     (S String 
        (name " " "."
         (str.++ S S) 
         (str.at S I)
         (str.replace S S S)   
         (str.substr S I I)))
      (I Int 
         (0 1 2 (+ I I) (- I I)
          (str.len S)
          (str.indexof S S I)))))

; Specifications to satisfy
(constraint (= (f "Nancy FreeHafer") "N.F."))
(constraint (= (f "Andrew Cencici") "A.C."))
(constraint (= (f "Jan Kotas") "J.K."))
(constraint (= (f "Mariya Sergienko") "M.S."))
\end{lstlisting}

%% file: codes/sygus_bv4.tex
\begin{lstlisting}[ tabsize=3, 
    basicstyle= \tt \footnotesize, 
    keywordstyle=\color{black}\bfseries, 
    commentstyle=\color{gray}, 
    escapeinside=``,
    language=Lisp,
    morekeywords = {set-logic, synth-fun, declare-var, constraint, check-synth, Int, Bool, BitVec},
    deletekeywords = {replace, min, max},
    numbers = none
    ]
; Determines what terms can appear
(set-logic BV)
; The function to synthesize
(synth-fun inv 
    ((s (BitVec 4)) (t (BitVec 4))) 
    (BitVec 4))
; Helper functions 
(define-fun min () (BitVec 4)
    (bvnot (bvlshr (bvnot #x0) #x1)))
(define-fun max () (BitVec 4) (bvnot min))
(define-fun l 
    ((s (BitVec 4)) (t (BitVec 4))) Bool
    (bvsgt (bvnot (inv s t)) t))
(define-fun SC 
    ((s (BitVec 4)) (t (BitVec 4))) Bool
    (distinct t max))

; Specifications to satisfy
; with universally quantified variables
(declare-var s (BitVec 4))
(declare-var t (BitVec 4))
(constraint (=> (SC s t) (l s t)))
\end{lstlisting}

%% file: codes/grammar_slia.tex
\[
\begin{array}{rl}
    \T{Start} ::= & \T{S}   \\
    \T{S} ::= & \T{name} \mid \T{" "} \mid \T{"."} \\
    \mid &  \T{str.++ S S} \mid \T{str.at S I} \\
    \mid & \T{str.replace S S S} \\
    \mid & \T{str.substr S I I} \\
    \T{I} ::= & \T{0} \mid \T{1} \mid \T{2} \mid \T{+ I I} \mid \T{- I I} \\
    \mid & \T{str.len S} \mid \T{str.indexof S S I} \\
\end{array}
\]

%% file: codes/grammar_bv4.tex
\[
\begin{array}{rl}
    \T{Start} ::= & \T{BV}   \\
    \T{BV} ::= & \T{s} \mid \T{t} \\
    \mid & \T{#x0} \mid \T{#x7} \mid \T{#x8} \\
    \mid & \T{bvneg BV} \mid \T{bvnot BV} \\
    \mid & \T{bvadd BV BV} \mid \T{bvsub BV BV} \\
    \mid & \T{bvand BV BV} \mid \T{bvlor BV BV} \\
    \mid & \T{bvlshl BV BV} \mid \T{bvlshr BV BV} \\
\end{array}
\]

%% file: related.tex
\section{Related Work}\label{sec:related}

\paragraph{Constrained Decoding}
Past work has extensively explored \textit{constrained decoding} algorithms, which modify the original decoding process of LLMs to ensure the output adheres to a user-specified regular \cite{melcer2024constrained, willard2023efficient} or context-free language \cite{lmql, dong2022codepad, geng2024grammarconstrained, poesia2022synchromesh, picard, shin2021constrained, stengel2023zero, ugare2024improving} in a discrete space.
Other works enforce hard output constraints using dynamic monitoring and verification methods~\cite{AgrawalKGLR23, li2024guiding, wang2023grammar} or by modifying beam search techniques to impose lexical constraints, which require specific keywords to appear in the generated text~\cite{anderson2016guided, hokamp2017lexically, hu2019improved, lu-etal-2022-neurologic, lu-etal-2021-neurologic, post2018fast}.
At a high level, these methods involve running the LLM decode in parallel with a monitoring scheme (e.g., parsing algorithms for CFGs) to identify which next tokens or beams can produce valid output sequences that meet the constraints. The decoder then masks out any tokens that would lead to invalid sequences, sampling only from the permissible ones.

To incorporate sequence-level soft semantic or contextual constraints, \citet{amini2024structured, kumar2022gradient, li2022diffusion, qin2022cold} have applied gradient-based sampling techniques that relax those constraints to differentiable ones, used them as classifiers to further guide the decoding process.
While these works guarantee that the decoded output meets the specified constraints (whether in the form of grammar, monitoring schemes, or differentiable functions), they often operate greedily and introduce bias into the output distribution in the way that has been discussed in this paper.
Depending on the application one considers, this problem may or may not affect downstream tasks, but as we have argued in this paper, the bias can be quite prominent and sometimes affect downstream performance.
Our adaptive decoding algorithm improves decoding over time by analyzing how previous samples led to nongrammaticaility.

\paragraph{Constraint-Aligned Decoding}
This paper formally defines the problem of aligning the output distribution of an LLM in the presence of a constraint.
We focus our attention on constraints expressed as grammars, but our definitions and algorithm apply to any constraint for which possible satisfaction (in our case grammaticality) can be evaluated in a left-to-right manner.

In some settings, one is interested in generating multiple outcomes with an LLM to approximate a distribution of interest~\cite{huang2024large, llm_sampling_renda_hopkins_2023}---e.g., to generate a random number or a set of good test cases for a program.
As we have shown, constrained decoding can heavily skew the LLMs distribution and result in biasing the model towards certain constraint-matching sequences.
While our work is at this point theoretical, now that the problem of aligning an LLM's distribution with constraints has been defined, we expect advances in how sampling is performed to quickly converge to better distributions faster (e.g., using beam search to quickly explore possible paths instead of just sampling).

\section{Conclusion}

We have introduced a new analysis of the ideal target for constrained sampling from an LLM using a grammar, which we call grammar-aligned decoding (GAD). We proposed a new algorithm for GAD which we call ASAp that iteratively builds better approximations to the critical re-weighting term required for GAD: the expected future grammaticality. We analyzed the convergence of our proposed algorithm and demonstrated its effectiveness in relation to existing grammar-constrained decoding techniques on a set of benchmark code generation tasks. We analyzed and evaluated our approach using constraints enforced by a context-free grammar; however, extensions of our approach might be applied to more general classes of constraints for LLM decoding.


While the primary goals of this work are to formalize the likelihood misalignment problem of existing grammar-constrained decoding approaches and to provide an initial solution with provable asymptotic guarantees, future work may explore faster-converging approaches, such as sampling multiple tokens simultaneously, to improve efficiency further. We hope this work lays a solid foundation for generating structured outputs from LLMs without distorting the original distribution, advancing the field toward more efficient, trustworthy, and constraint-aligned approaches in LLM-driven generation.

%% file: appendix.tex

\section{Hardware and Software}\label{app:hardware}

Our experiments are conducted on 4 NVIDIA RTX A6000 GPUs and 4 NVIDIA A100 GPUs. Our implementation is based on Python 3.10 and PyTorch 2.1.2.

\section{Hyperparameters}\label{app:hyperparams}
The hyperparameters discussed in this paper pertain to the decoding strategy of language models. As we aim to investigate the LM's original distribution, we set Top-P at $1.0$, Temperature at $1.0$, and Top-K at $0$ to consider the complete token vocabulary.

\section{Model Checkpoint}\label{app:models}
We use the Mistral-7B model checkpoint provided by Hugging Face:  \url{https://huggingface.co/mistralai/Mistral-7B-Instruct-v0.2}.

\section{Experimental Details}\label{app:exp_detail}

\subsection{SLIA and INV-BV}

\paragraph{Prompt Construction}
For both families of benchmarks, our prompts adopt standard in-context learning format which consist of 3 in-context examples of the form (specification, solution) and ask the model to provide the solution for the last example. A concrete example would be
\begin{verbatim}
You are an expert in program synthesis. 
You are tasked with solving a Syntax-Guided Synthesis (SyGuS) problem. 
Your goal is to output a function that should produce outputs that satisfy 
a series of constraints when given specific inputs.

Question:
(set-logic BV)

(synth-fun inv ((s (BitVec 4)) (t (BitVec 4))) (BitVec 4))

(declare-var s (BitVec 4))
(declare-var t (BitVec 4))
(define-fun udivtotal ((a (BitVec 4)) (b (BitVec 4))) (BitVec 4)
    (ite (= b #x0) #xF (bvudiv a b)))
(define-fun uremtotal ((a (BitVec 4)) (b (BitVec 4))) (BitVec 4)
    (ite (= b #x0) a (bvurem a b)))
(define-fun min () (BitVec 4)
    (bvnot (bvlshr (bvnot #x0) #x1)))
(define-fun max () (BitVec 4)
    (bvnot min))
(define-fun l ((s (BitVec 4)) (t (BitVec 4))) Bool
    (bvsle (bvlshr s (inv s t)) t))
(define-fun SC ((s (BitVec 4)) (t (BitVec 4))) Bool
    (or (bvult t min) (bvsge t s)))
(constraint (=> (SC s t) (l s t)))

(check-synth)
Solution:
(define-fun inv ((s (BitVec 4)) (t (BitVec 4))) (BitVec 4) (bvnot (bvor s #b0111)))

... (2 more examples)

Question:
(set-logic BV)

(synth-fun inv ((s (BitVec 4)) (t (BitVec 4))) (BitVec 4))

(declare-var s (BitVec 4))
(declare-var t (BitVec 4))
(define-fun udivtotal ((a (BitVec 4)) (b (BitVec 4))) (BitVec 4)
    (ite (= b #x0) #xF (bvudiv a b)))
(define-fun uremtotal ((a (BitVec 4)) (b (BitVec 4))) (BitVec 4)
    (ite (= b #x0) a (bvurem a b)))
(define-fun min () (BitVec 4)
    (bvnot (bvlshr (bvnot #x0) #x1)))
(define-fun max () (BitVec 4)
    (bvnot min))
(define-fun l ((s (BitVec 4)) (t (BitVec 4))) Bool
    (bvsgt (bvnot (inv s t)) t))
(define-fun SC ((s (BitVec 4)) (t (BitVec 4))) Bool
    (distinct t max))
(constraint (=> (SC s t) (l s t)))

(check-synth)
Solution:
\end{verbatim}

\paragraph{Grammar Constraint}
While most \sygus problems contain grammar constraints, some problems have grammars implicitly defined by the theory. We explicitly converted the grammar constraint of the problem into EBNF format for constrained-decoding. The example for the last example would be
\begin{verbatim}
root ::= "(define-fun inv ((s (BitVec 4)) (t (BitVec 4))) (BitVec 4) " Start ")"
Start ::= "s" | "t" | "#x0" | "#x8" | "#x7" 
        | "(" "bvneg" " " Start ")" | "(" "bvnot" " " Start ")" 
        | "(" "bvadd" " " Start " " Start ")" | "(" "bvsub" " " Start " " Start ")" 
        | "(" "bvand" " " Start " " Start ")" | "(" "bvlshr" " " Start " " Start ")" 
        | "(" "bvor" " " Start " " Start ")" | "(" "bvshl" " " Start " " Start ")"
\end{verbatim}

\subsection{Constituency Parsing}

For Constituency parsing task, our prompts consist of 8 in-context examples of the form. A concrete example would be
\begin{verbatim}
Perform constituency parsing on the provided sentences in accordance with the Penn TreeBank 
annotation guidelines. Fill in the last mapping.

Ad Notes
->
[ ( NP-HLN ( NN Ad ) ( NNS Notes ) ) ]

The market crumbled
->
[ ( S ( NP-SBJ ( DT The ) ( NN market ) ) ( VP ( VBD crumbled ) ) ) ]

I felt betrayed he later said
->
[ ( S ( S-TPC-1 ( NP-SBJ ( PRP I ) ) ( VP ( VBD felt ) ( ADJP-PRD ( VBN betrayed ) ) ) ) 
( NP-SBJ ( PRP he ) ) ( ADVP-TMP ( RB later ) ) ( VP ( VBD said ) ) ) ]

Friday October 13 1989
->
[ ( NP ( NNP Friday ) ( NNP October ) ( CD 13 ) ( CD 1989 ) ) ]

The Arabs had merely oil
->
[ ( S ( NP-SBJ ( DT The ) ( NNPS Arabs ) ) ( VP ( VBD had ) 
( NP ( RB merely ) ( NN oil ) ) ) ) ]

Energy
->
[ ( NP-HLN ( NN Energy ) ) ]

Some U.S. entrepreneurs operate on a smaller scale
->
[ ( S ( NP-SBJ ( DT Some ) ( NNP U.S. ) ( NNS entrepreneurs ) ) ( VP ( VBP operate ) 
( PP-MNR ( IN on ) ( NP ( DT a ) ( JJR smaller ) ( NN scale ) ) ) ) ) ]

Knowledgeware Inc.
->
[ ( NP-HLN ( NNP Knowledgeware ) ( NNP Inc. ) ) ]

They are more sophisticated this time
->
\end{verbatim}

\paragraph{Grammar Constraint}
For the constituency parsing (CP) task we used the grammar provided in prior GCD work~\cite{geng2024grammarconstrained}.
The grammar is too large to attach, but it is used to help the model produce well-parenthesized parse trees and ensure that all words in a given English sentence appear in left-to-right order.

\newpage
\section{Detailed Experimental Results}
\label{app:detailed-results}
We provide additional plots and experimental data.
\subsection{Plots}
\label{app:plots}

Figures~8--22 provide the KL divergence and expectation results for the SLIA benchmarks.
Figures~23--37 provide the KL divergence and expectation results for the INV-BV benchmarks.
Figures~38--43 provide the KL divergence and expectation results for the INV-BV benchmarks.

\begin{figure}[t]
\begin{minipage}{0.49\textwidth}
\begin{subfigure}[valign=t]{0.49\textwidth}
\includegraphics[width=\textwidth]{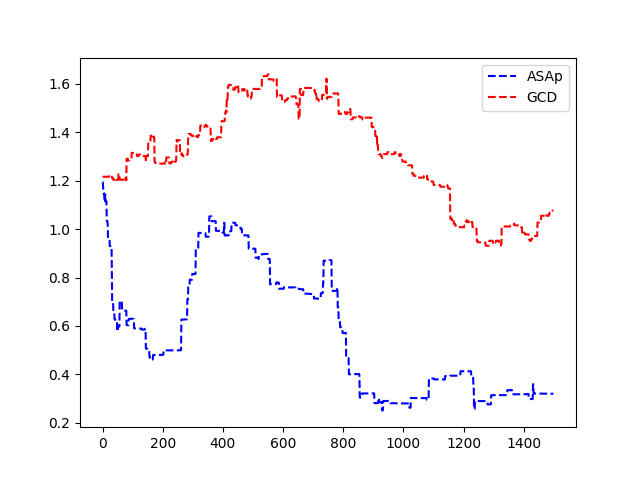}
\caption{KL divergences}
\end{subfigure}
\begin{subfigure}[valign=t]{0.49\textwidth}
\includegraphics[width=\textwidth]{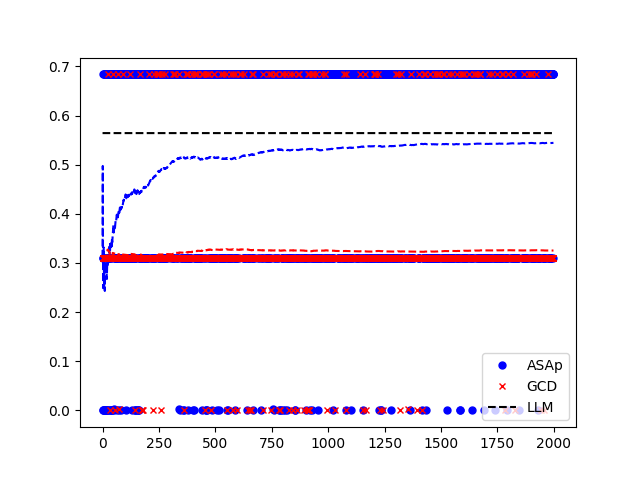}
\caption{Expectations}
\end{subfigure}
\caption{\texttt{SLIA/dr-name}}
\end{minipage}
\hfill
\begin{minipage}{0.49\textwidth}
\begin{subfigure}[valign=t]{0.49\textwidth}
\includegraphics[width=\textwidth]{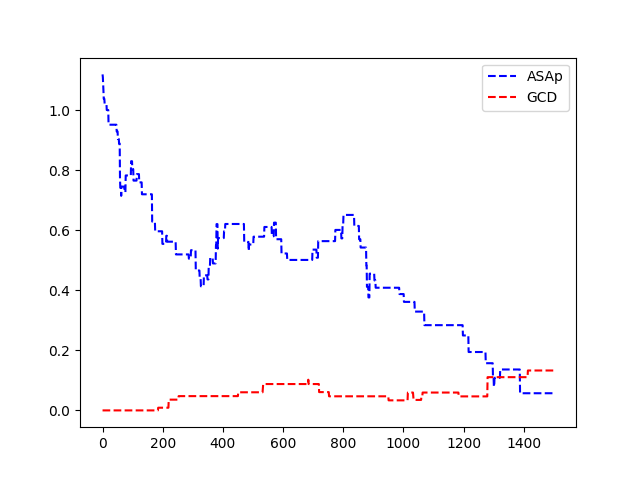}
\caption{KL divergences}
\end{subfigure}
\begin{subfigure}[valign=t]{0.49\textwidth}
\includegraphics[width=\textwidth]{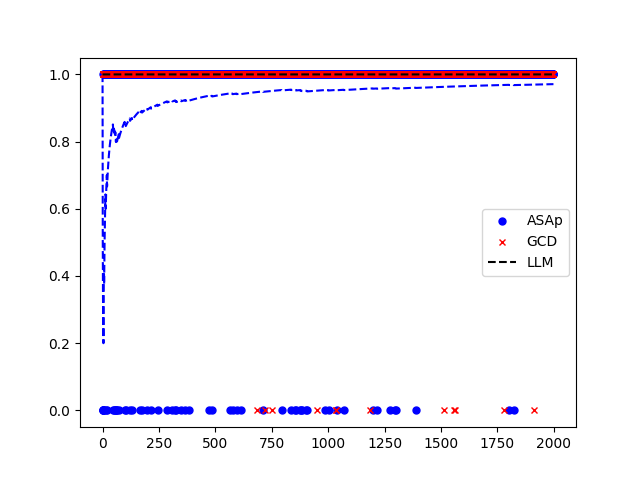}
\caption{Expectations}
\end{subfigure}
\caption{\texttt{SLIA/firstname\_small}}
\end{minipage}

\begin{minipage}{0.49\textwidth}
\begin{subfigure}[valign=t]{0.49\textwidth}
\includegraphics[width=\textwidth]{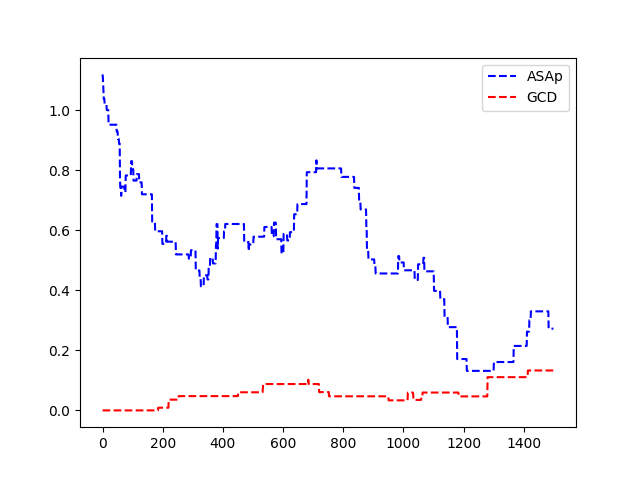}
\caption{KL divergences}
\end{subfigure}
\begin{subfigure}[valign=t]{0.49\textwidth}
\includegraphics[width=\textwidth]{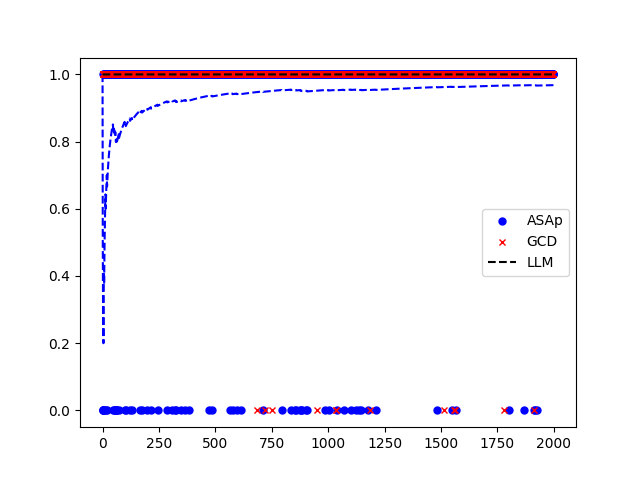}
\caption{Expectations}
\end{subfigure}
\caption{\texttt{SLIA/firstname}}
\end{minipage}
\hfill
\begin{minipage}{0.49\textwidth}
\begin{subfigure}[valign=t]{0.49\textwidth}
\includegraphics[width=\textwidth]{figs/SLIA/kl/initials_small.png}
\caption{KL divergences}
\end{subfigure}
\begin{subfigure}[valign=t]{0.49\textwidth}
\includegraphics[width=\textwidth]{figs/SLIA/prob/initials_small.png}
\caption{Expectations}
\end{subfigure}
\caption{\texttt{SLIA/initials\_small}}
\end{minipage}

\begin{minipage}{0.49\textwidth}
\begin{subfigure}[valign=t]{0.49\textwidth}
\includegraphics[width=\textwidth]{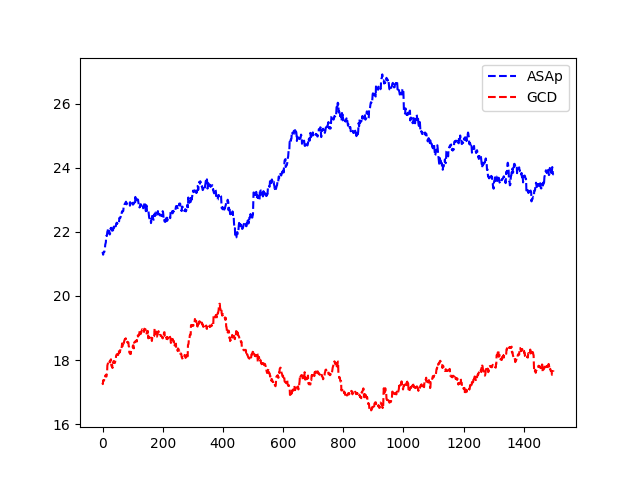}
\caption{KL divergences}
\end{subfigure}
\begin{subfigure}[valign=t]{0.49\textwidth}
\includegraphics[width=\textwidth]{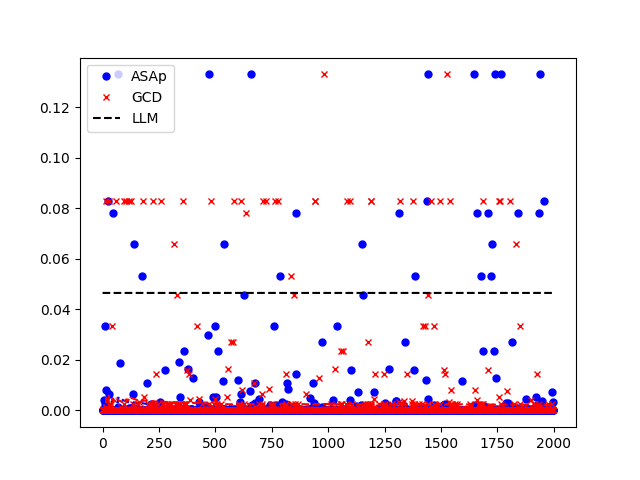}
\caption{Expectations}
\end{subfigure}
\caption{\texttt{SLIA/initials-long-repeat}}
\end{minipage}
\hfill
\begin{minipage}{0.49\textwidth}
\begin{subfigure}[valign=t]{0.49\textwidth}
\includegraphics[width=\textwidth]{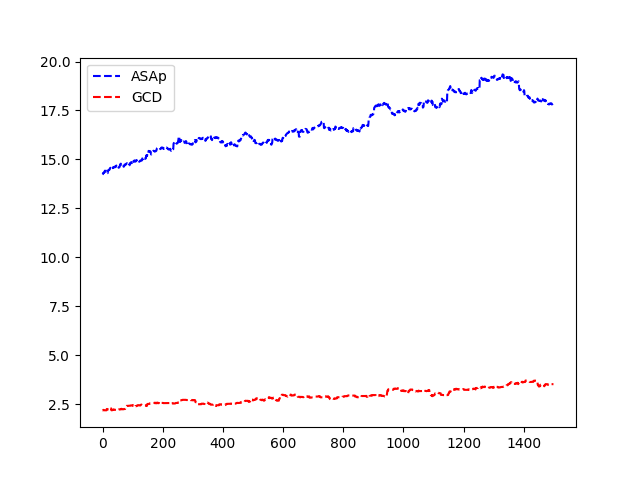}
\caption{KL divergences}
\end{subfigure}
\begin{subfigure}[valign=t]{0.49\textwidth}
\includegraphics[width=\textwidth]{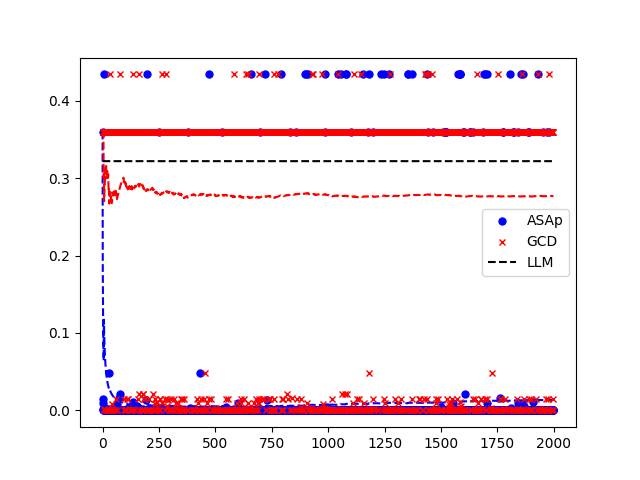}
\caption{Expectations}
\end{subfigure}
\caption{\texttt{SLIA/lastname}}
\end{minipage}

\begin{minipage}{0.49\textwidth}
\begin{subfigure}[valign=t]{0.49\textwidth}
\includegraphics[width=\textwidth]{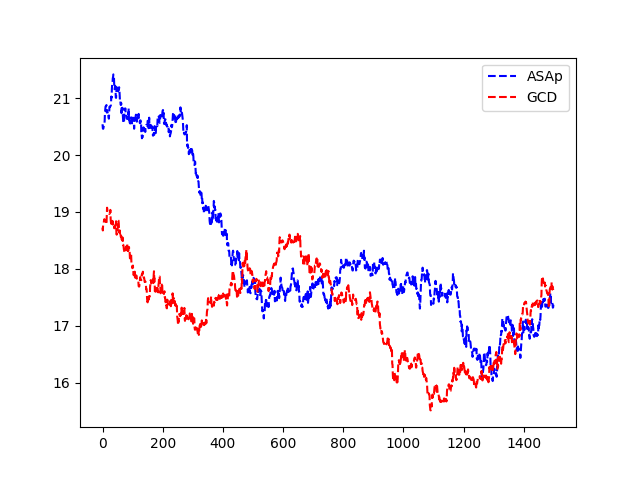}
\caption{KL divergences}
\end{subfigure}
\begin{subfigure}[valign=t]{0.49\textwidth}
\includegraphics[width=\textwidth]{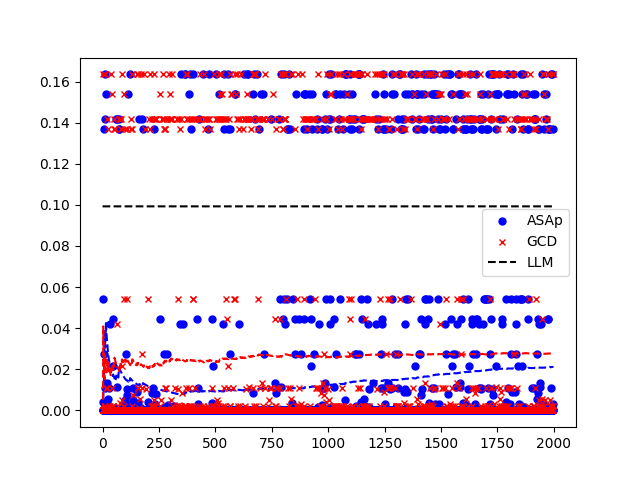}
\caption{Expectations}
\end{subfigure}
\caption{\texttt{SLIA/name-combine-2\_short}}
\end{minipage}
\hfill
\begin{minipage}{0.49\textwidth}
\begin{subfigure}[valign=t]{0.49\textwidth}
\includegraphics[width=\textwidth]{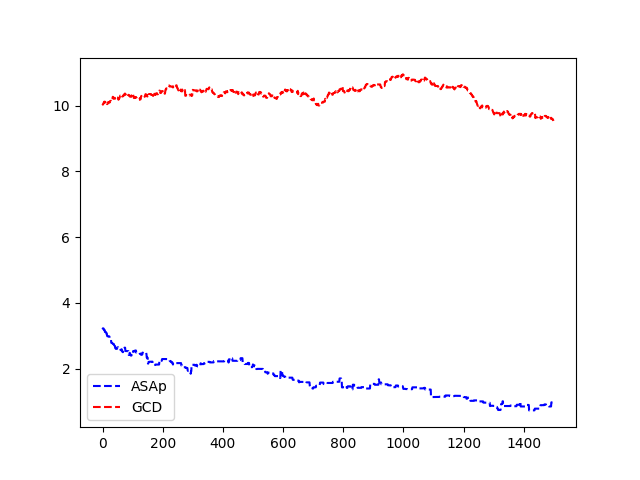}
\caption{KL divergences}
\end{subfigure}
\begin{subfigure}[valign=t]{0.49\textwidth}
\includegraphics[width=\textwidth]{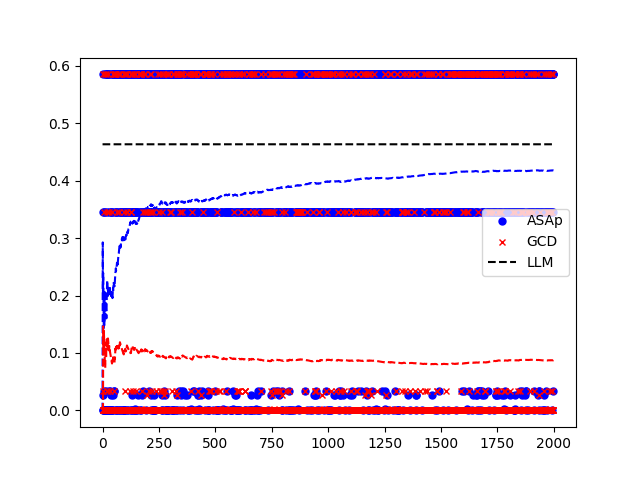}
\caption{Expectations}
\end{subfigure}
\caption{\texttt{SLIA/name-combine-2-long-repeat}}
\end{minipage}

\begin{minipage}{0.49\textwidth}
\begin{subfigure}[valign=t]{0.49\textwidth}
\includegraphics[width=\textwidth]{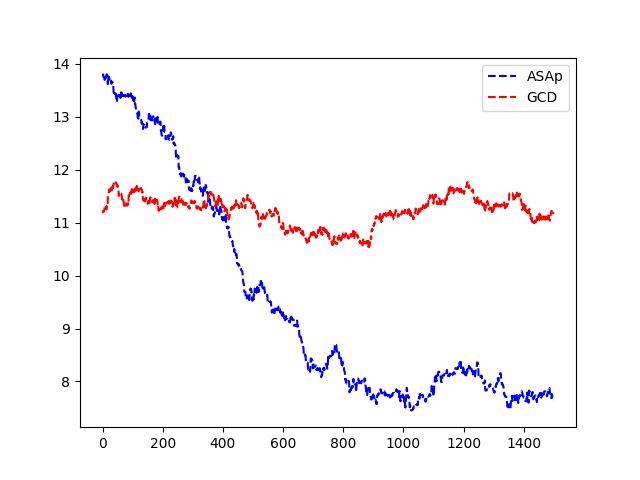}
\caption{KL divergences}
\end{subfigure}
\begin{subfigure}[valign=t]{0.49\textwidth}
\includegraphics[width=\textwidth]{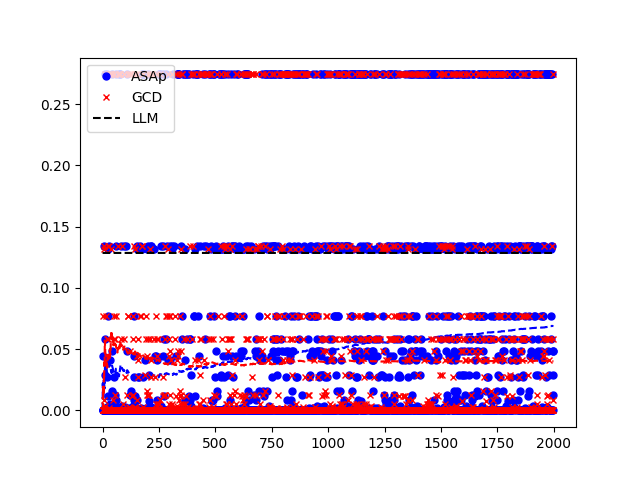}
\caption{Expectations}
\end{subfigure}
\caption{\texttt{SLIA/name-combine-4\_short}}
\end{minipage}
\hfill
\begin{minipage}{0.49\textwidth}
\begin{subfigure}[valign=t]{0.49\textwidth}
\includegraphics[width=\textwidth]{figs/SLIA/kl/name-combine-4-long.png}
\caption{KL divergences}
\end{subfigure}
\begin{subfigure}[valign=t]{0.49\textwidth}
\includegraphics[width=\textwidth]{figs/SLIA/prob/name-combine-4-long.png}
\caption{Expectations}
\end{subfigure}
\caption{\texttt{SLIA/name-combine-4-long}}
\end{minipage}
\end{figure}

\begin{figure}[t]
\begin{minipage}{0.49\textwidth}
\begin{subfigure}[valign=t]{0.49\textwidth}
\includegraphics[width=\textwidth]{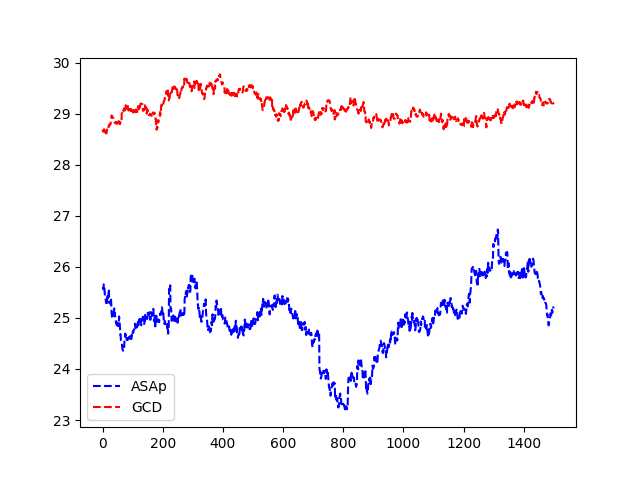}
\caption{KL divergences}
\end{subfigure}
\begin{subfigure}[valign=t]{0.49\textwidth}
\includegraphics[width=\textwidth]{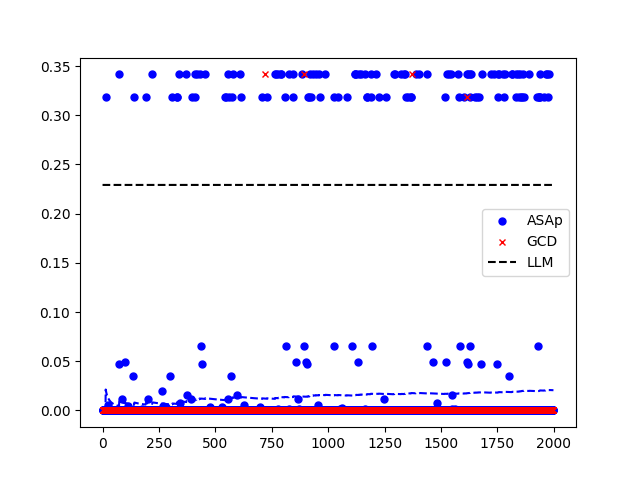}
\caption{Expectations}
\end{subfigure}
\caption{\texttt{SLIA/phone-3-long}}
\end{minipage}
\hfill
\begin{minipage}{0.49\textwidth}
\begin{subfigure}[valign=t]{0.49\textwidth}
\includegraphics[width=\textwidth]{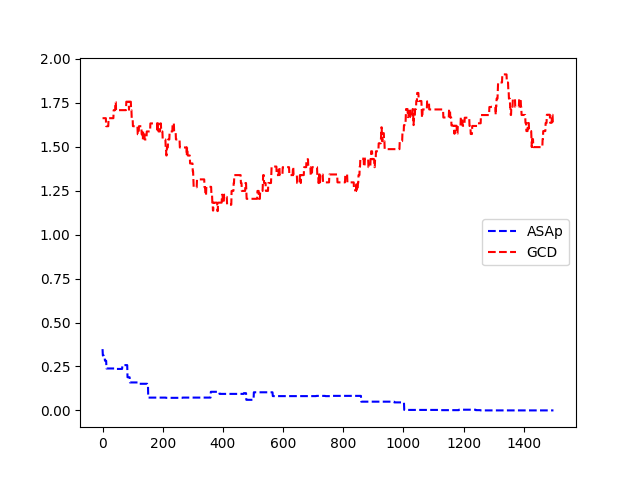}
\caption{KL divergences}
\end{subfigure}
\begin{subfigure}[valign=t]{0.49\textwidth}
\includegraphics[width=\textwidth]{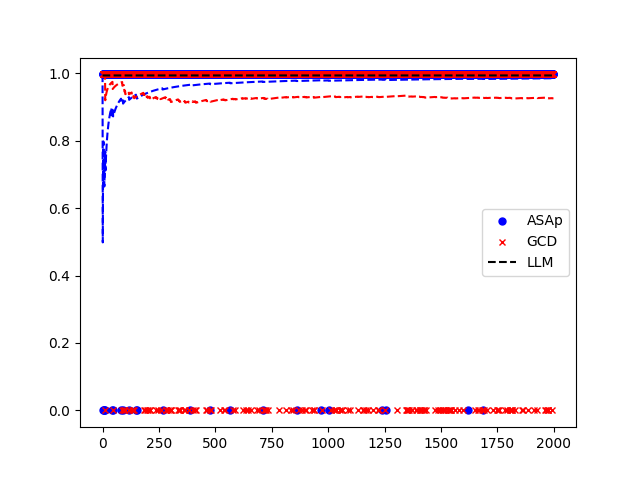}
\caption{Expectations}
\end{subfigure}
\caption{\texttt{SLIA/reverse-name-long}}
\end{minipage}
\end{figure}

\begin{figure}[t]
\begin{minipage}{0.49\textwidth}
\begin{subfigure}[valign=t]{0.49\textwidth}
\includegraphics[width=\textwidth]{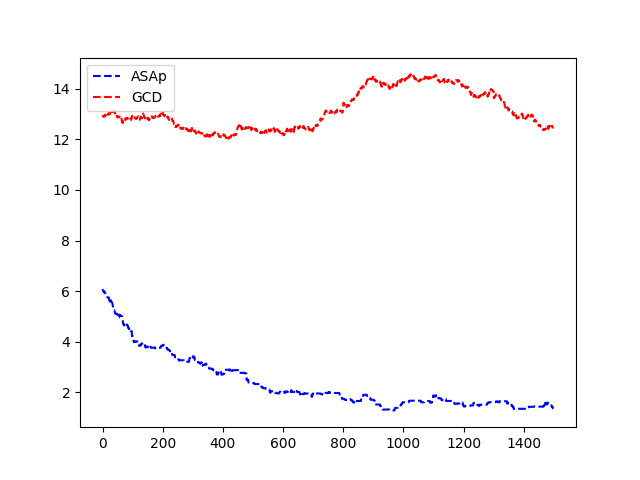}
\caption{KL divergences}
\end{subfigure}
\begin{subfigure}[valign=t]{0.49\textwidth}
\includegraphics[width=\textwidth]{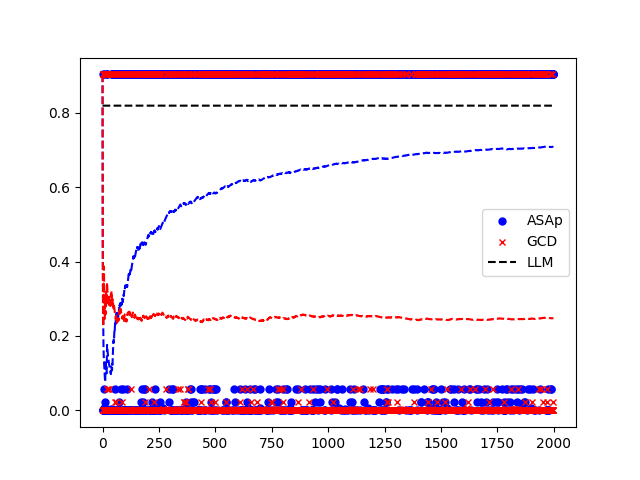}
\caption{Expectations}
\end{subfigure}
\caption{\texttt{SLIA/univ\_1\_short}}
\end{minipage}
\hfill
\begin{minipage}{0.49\textwidth}
\begin{subfigure}[valign=t]{0.49\textwidth}
\includegraphics[width=\textwidth]{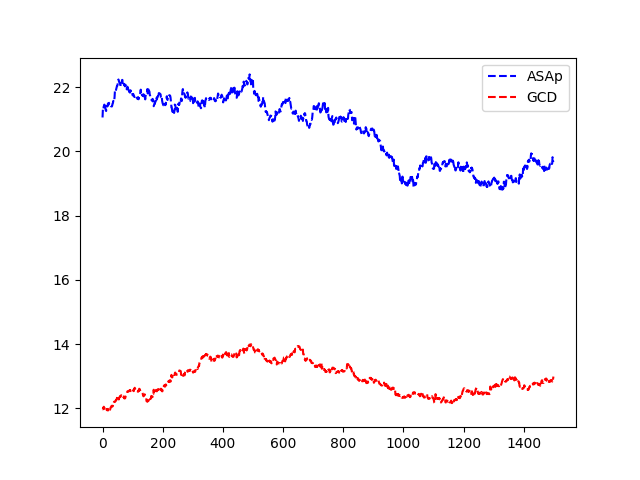}
\caption{KL divergences}
\end{subfigure}
\begin{subfigure}[valign=t]{0.49\textwidth}
\includegraphics[width=\textwidth]{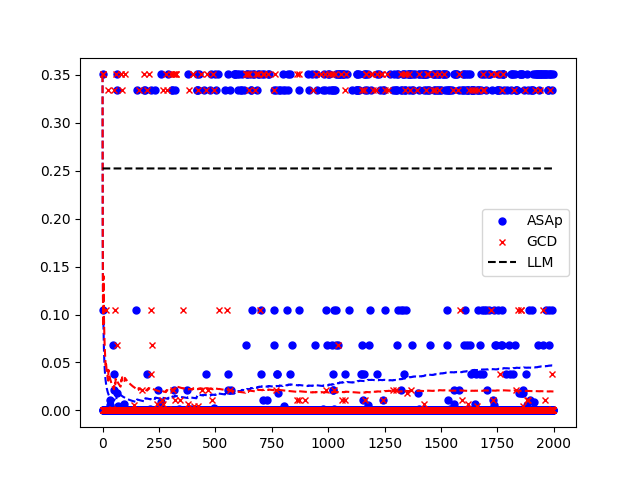}
\caption{Expectations}
\end{subfigure}
\caption{\texttt{SLIA/univ\_1}}
\end{minipage}

\begin{minipage}{0.49\textwidth}
\begin{subfigure}[valign=t]{0.49\textwidth}
\includegraphics[width=\textwidth]{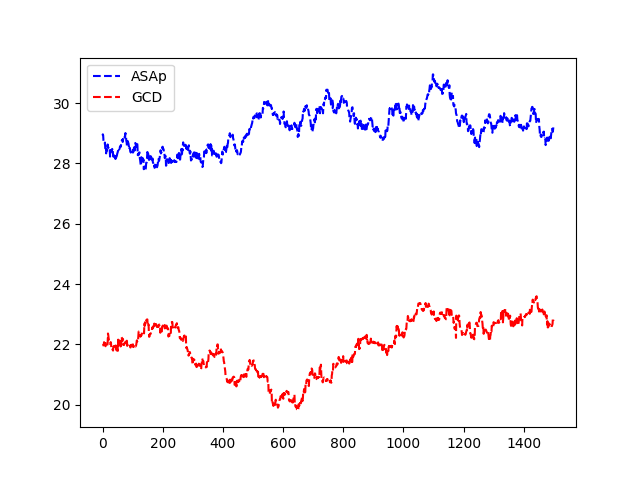}
\caption{KL divergences}
\end{subfigure}
\begin{subfigure}[valign=t]{0.49\textwidth}
\includegraphics[width=\textwidth]{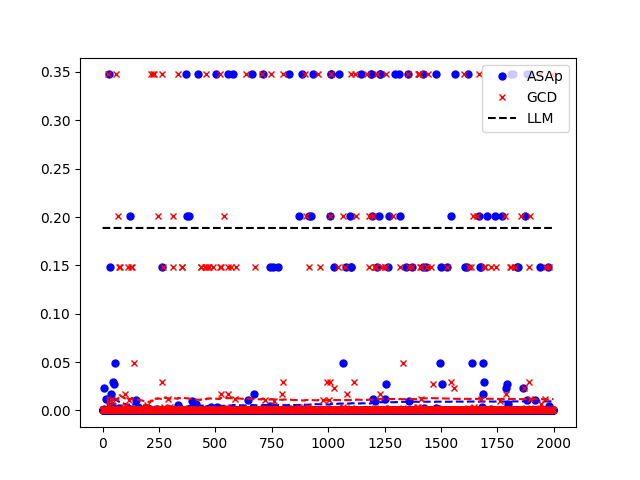}
\caption{Expectations}
\end{subfigure}
\caption{\texttt{SLIA/univ\_2\_short}}
\end{minipage}
\hfill
\begin{minipage}{0.49\textwidth}
\begin{subfigure}[valign=t]{0.49\textwidth}
\includegraphics[width=\textwidth]{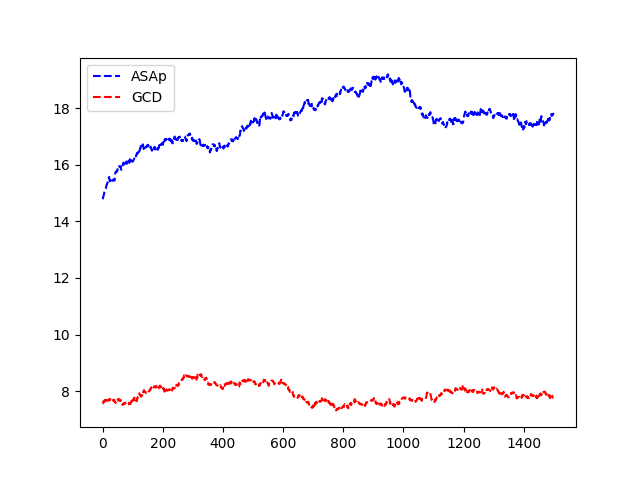}
\caption{KL divergences}
\end{subfigure}
\begin{subfigure}[valign=t]{0.49\textwidth}
\includegraphics[width=\textwidth]{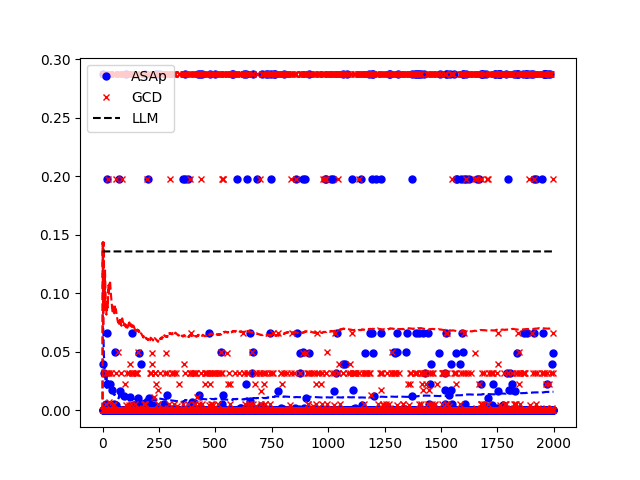}
\caption{Expectations}
\end{subfigure}
\caption{\texttt{INV-BV/find\_inv\_bvsge\_bvlshr1\_4bit}}
\end{minipage}

\begin{minipage}{0.49\textwidth}
\begin{subfigure}[valign=t]{0.49\textwidth}
\includegraphics[width=\textwidth]{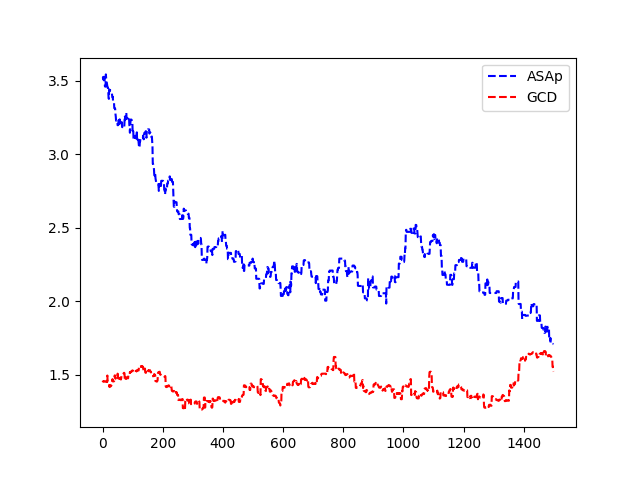}
\caption{KL divergences}
\end{subfigure}
\begin{subfigure}[valign=t]{0.49\textwidth}
\includegraphics[width=\textwidth]{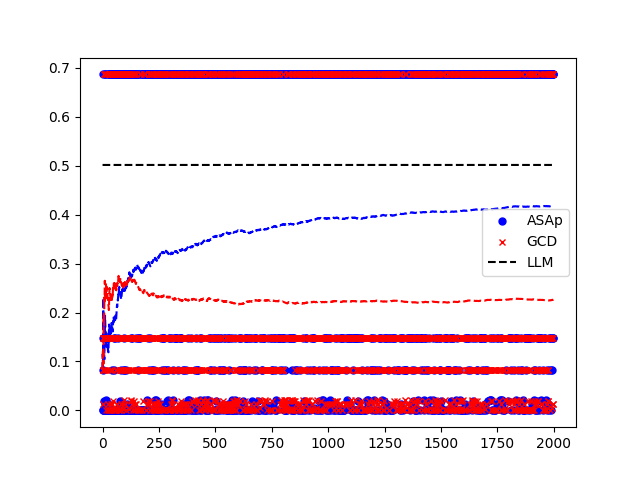}
\caption{Expectations}
\end{subfigure}
\caption{\texttt{INV-BV/find\_inv\_bvsge\_bvneg\_4bit}}
\end{minipage}
\begin{minipage}{0.49\textwidth}
\begin{subfigure}[valign=t]{0.49\textwidth}
\includegraphics[width=\textwidth]{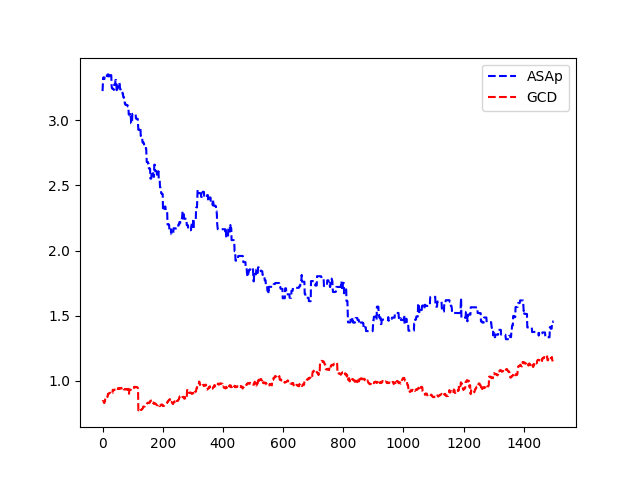}
\caption{KL divergences}
\end{subfigure}
\begin{subfigure}[valign=t]{0.49\textwidth}
\includegraphics[width=\textwidth]{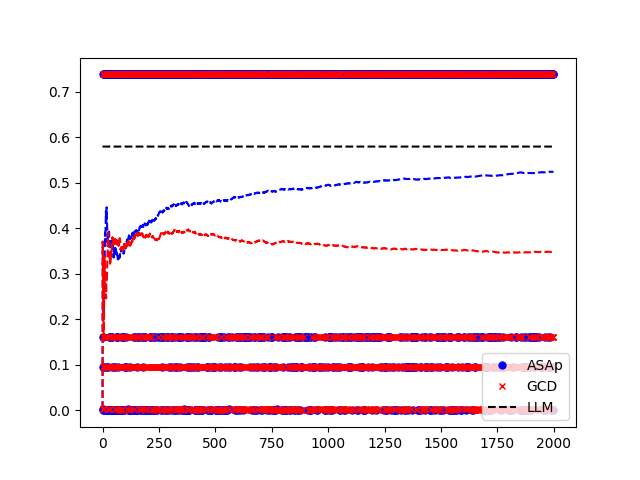}
\caption{Expectations}
\end{subfigure}
\caption{\texttt{INV-BV/find\_inv\_bvsge\_bvnot\_4bit}}
\end{minipage}
\hfill 
\begin{minipage}{0.49\textwidth}
\begin{subfigure}[valign=t]{0.49\textwidth}
\includegraphics[width=\textwidth]{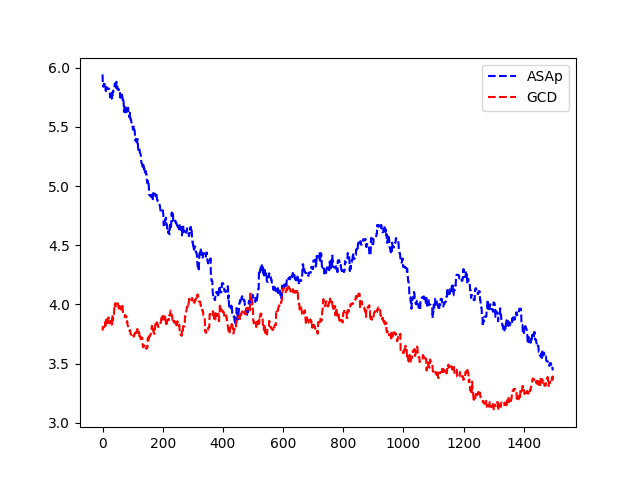}
\caption{KL divergences}
\end{subfigure}
\begin{subfigure}[valign=t]{0.49\textwidth}
\includegraphics[width=\textwidth]{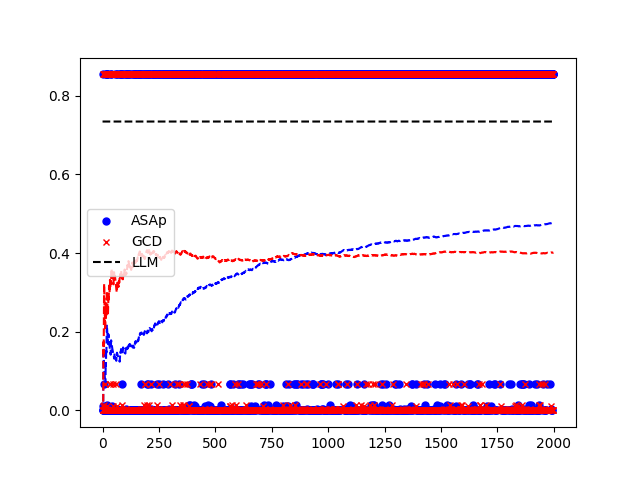}
\caption{Expectations}
\end{subfigure}
\caption{\texttt{INV-BV/find\_inv\_bvsgt\_bvor\_4bit}}
\end{minipage}
\hfill
\begin{minipage}{0.49\textwidth}
\begin{subfigure}[valign=t]{0.49\textwidth}
\includegraphics[width=\textwidth]{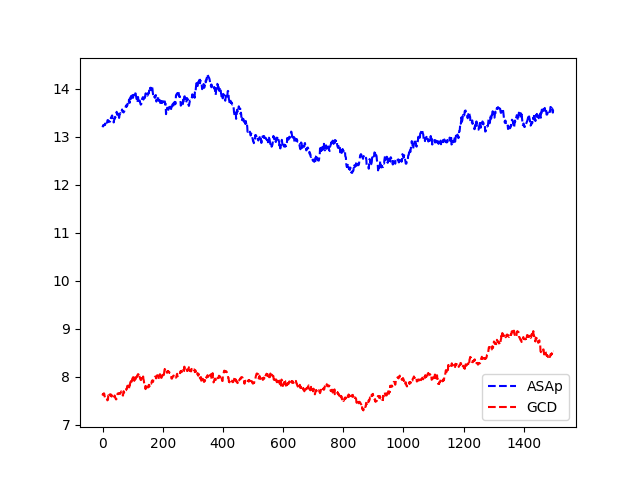}
\caption{KL divergences}
\end{subfigure}
\begin{subfigure}[valign=t]{0.49\textwidth}
\includegraphics[width=\textwidth]{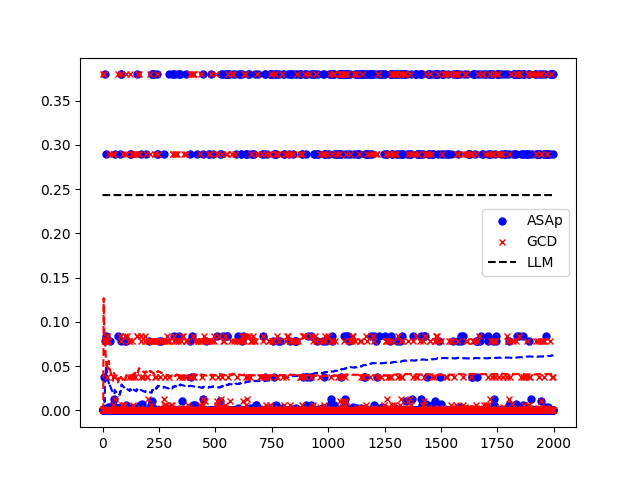}
\caption{Expectations}
\end{subfigure}
\caption{\texttt{INV-BV/find\_inv\_bvugt\_bvashr0\_4bit}}
\end{minipage}
\end{figure}

\begin{figure}[t]
\begin{minipage}{0.49\textwidth}
\begin{subfigure}[valign=t]{0.49\textwidth}
\includegraphics[width=\textwidth]{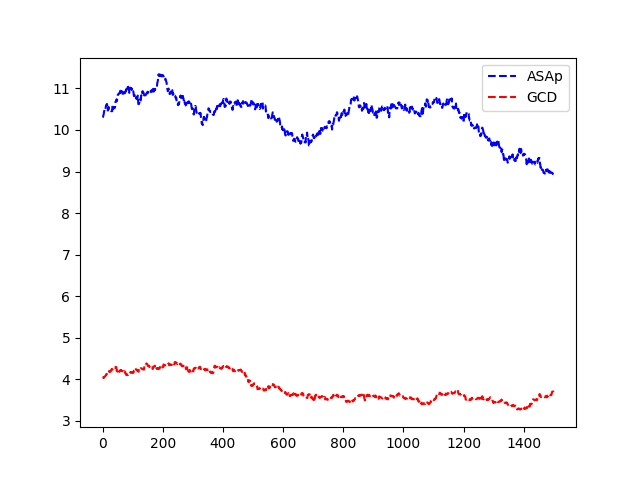}
\caption{KL divergences}
\end{subfigure}
\begin{subfigure}[valign=t]{0.49\textwidth}
\includegraphics[width=\textwidth]{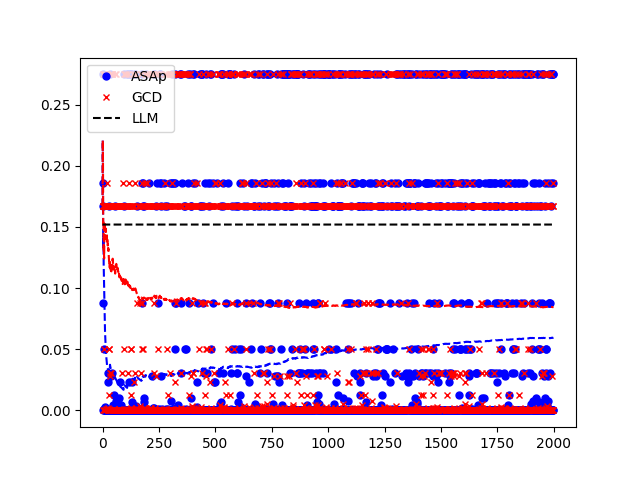}
\caption{Expectations}
\end{subfigure}
\caption{\texttt{INV-BV/find\_inv\_bvugt\_bvneg\_4bit}}
\end{minipage}
\hfill
\begin{minipage}{0.49\textwidth}
\begin{subfigure}[valign=t]{0.49\textwidth}
\includegraphics[width=\textwidth]{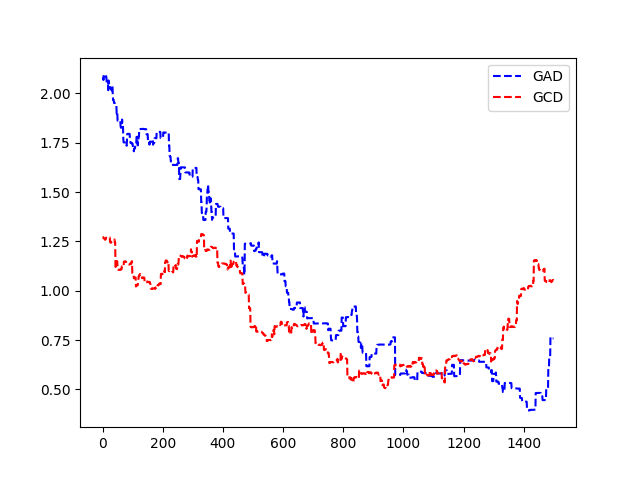}
\caption{KL divergences}
\end{subfigure}
\begin{subfigure}[valign=t]{0.49\textwidth}
\includegraphics[width=\textwidth]{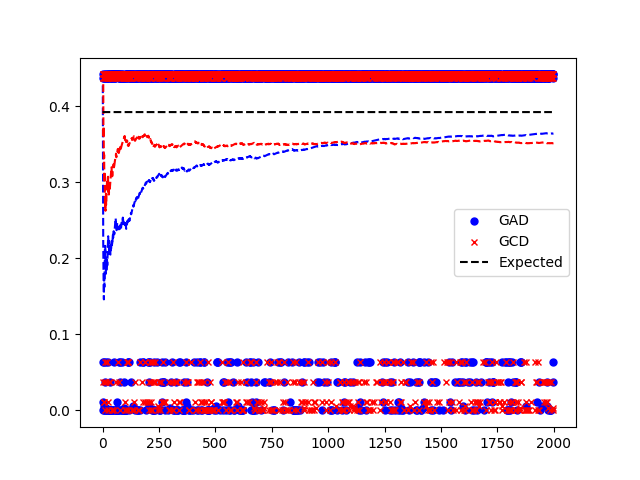}
\caption{Expectations}
\end{subfigure}
\caption{\texttt{INV-BV/find\_inv\_bvule\_bvurem0\_4bit}}
\end{minipage}

\begin{minipage}{0.49\textwidth}
\begin{subfigure}[valign=t]{0.49\textwidth}
\includegraphics[width=\textwidth]{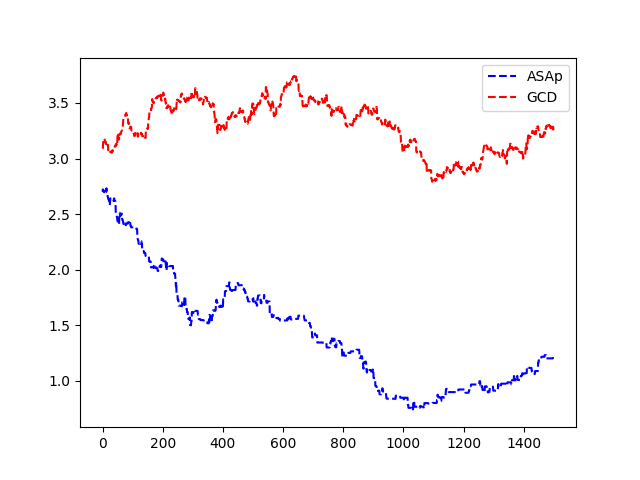}
\caption{KL divergences}
\end{subfigure}
\begin{subfigure}[valign=t]{0.49\textwidth}
\includegraphics[width=\textwidth]{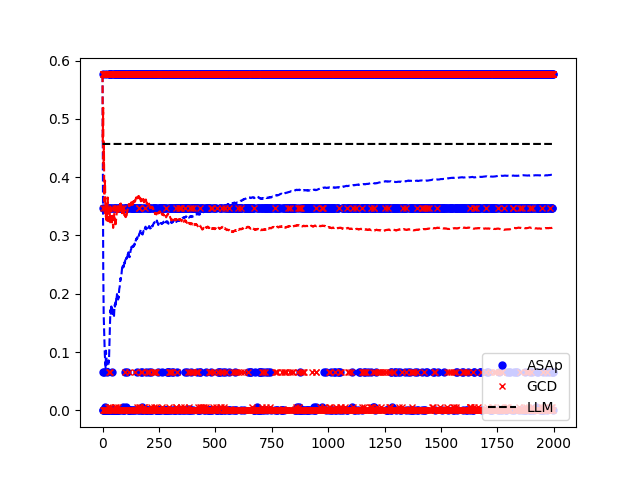}
\caption{Expectations}
\end{subfigure}
\caption{\texttt{INV-BV/find\_inv\_bvule\_bvurem1\_4bit}}
\end{minipage}
\hfill
\begin{minipage}{0.49\textwidth}
\begin{subfigure}[valign=t]{0.49\textwidth}
\includegraphics[width=\textwidth]{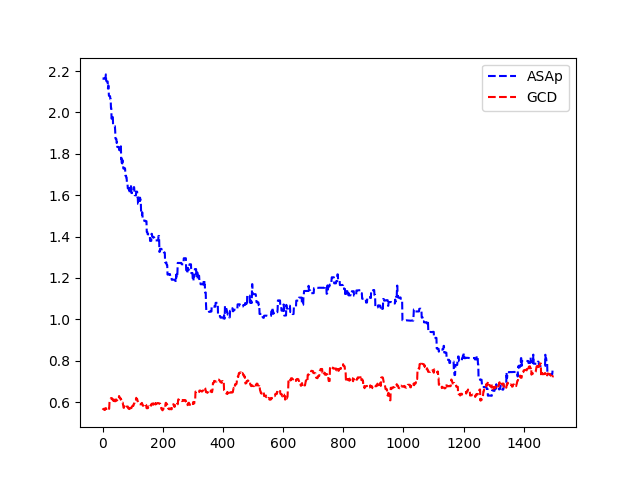}
\caption{KL divergences}
\end{subfigure}
\begin{subfigure}[valign=t]{0.49\textwidth}
\includegraphics[width=\textwidth]{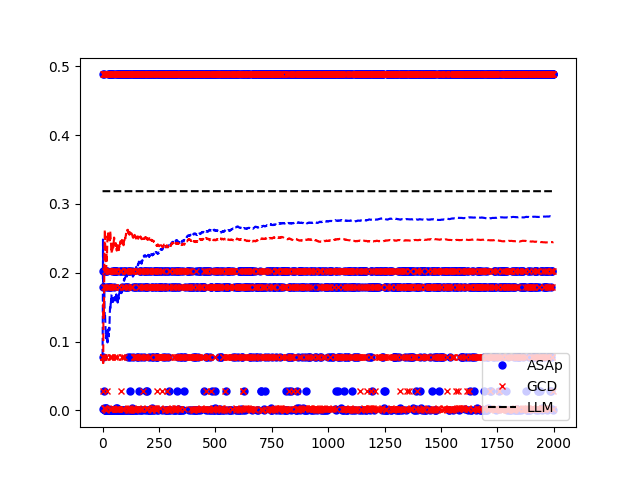}
\caption{Expectations}
\end{subfigure}
\caption{\texttt{INV-BV/find\_inv\_eq\_bvand\_4bit}}
\end{minipage}
\end{figure}

\begin{figure}[t]
\begin{minipage}{0.49\textwidth}
\begin{subfigure}[valign=t]{0.49\textwidth}
\includegraphics[width=\textwidth]{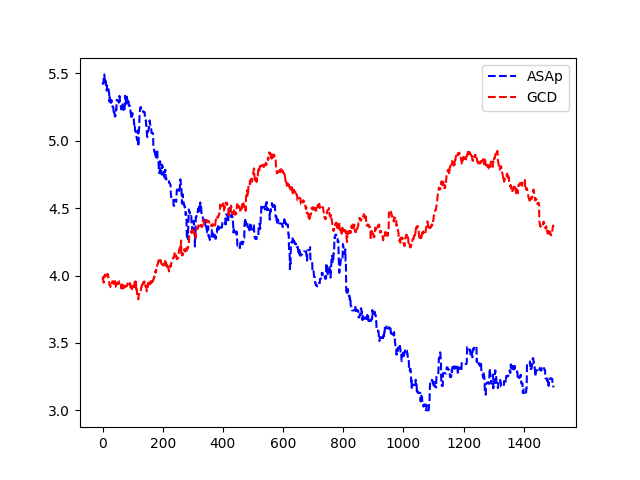}
\caption{KL divergences}
\end{subfigure}
\begin{subfigure}[valign=t]{0.49\textwidth}
\includegraphics[width=\textwidth]{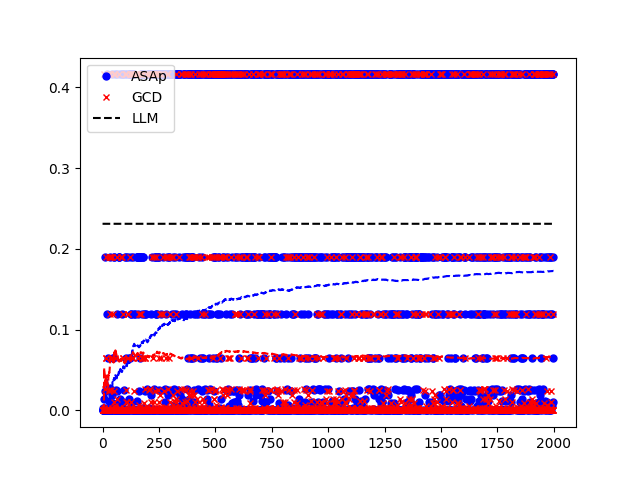}
\caption{Expectations}
\end{subfigure}
\caption{\texttt{INV-BV/find\_inv\_eq\_bvlshr0\_4bit}}
\end{minipage}
\hfill
\begin{minipage}{0.49\textwidth}
\begin{subfigure}[valign=t]{0.49\textwidth}
\includegraphics[width=\textwidth]{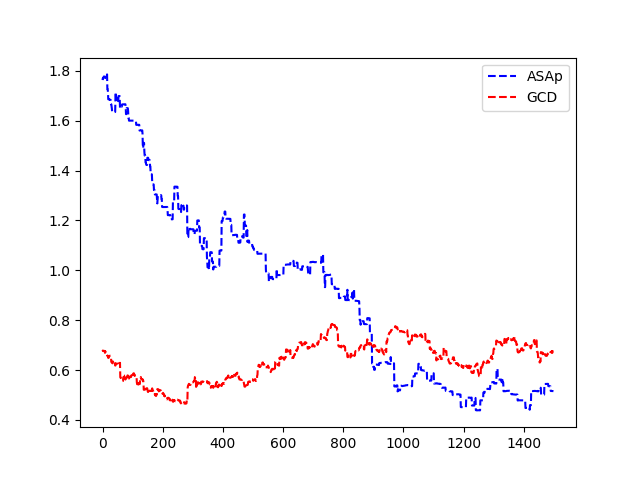}
\caption{KL divergences}
\end{subfigure}
\begin{subfigure}[valign=t]{0.49\textwidth}
\includegraphics[width=\textwidth]{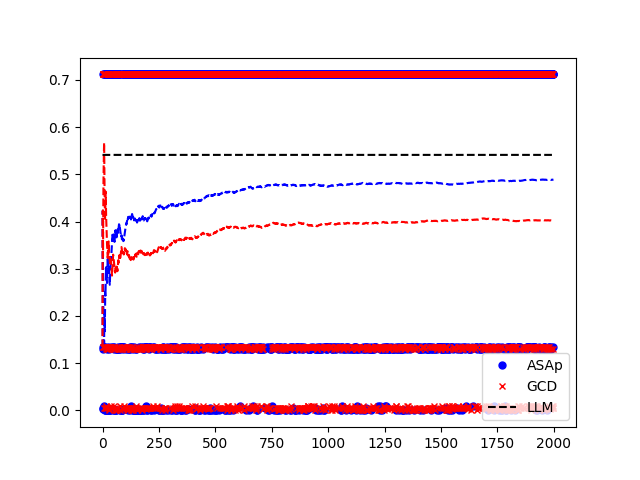}
\caption{Expectations}
\end{subfigure}
\caption{\texttt{INV-BV/find\_inv\_ne\_bvneg\_4bit}}
\end{minipage}

\begin{minipage}{0.49\textwidth}
\begin{subfigure}[valign=t]{0.49\textwidth}
\includegraphics[width=\textwidth]{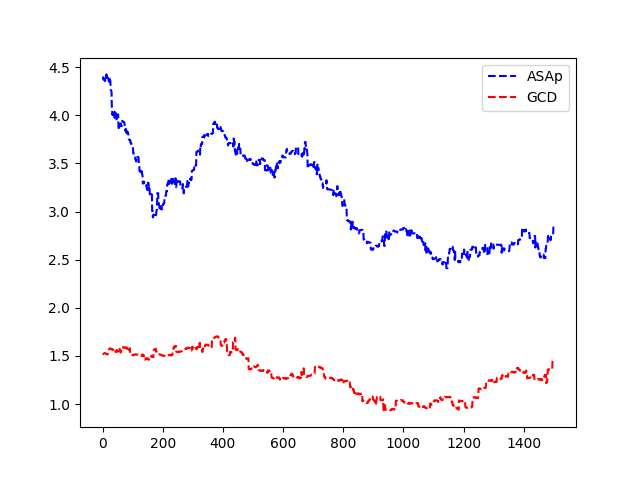}
\caption{KL divergences}
\end{subfigure}
\begin{subfigure}[valign=t]{0.49\textwidth}
\includegraphics[width=\textwidth]{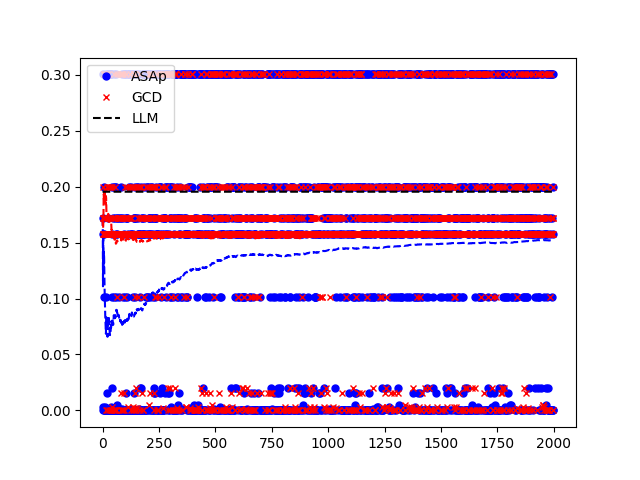}
\caption{Expectations}
\end{subfigure}
\caption{\texttt{INV-BV/find\_inv\_ne\_bvudiv0\_4bit}}
\end{minipage}
\hfill
\begin{minipage}{0.49\textwidth}
\begin{subfigure}[valign=t]{0.49\textwidth}
\includegraphics[width=\textwidth]{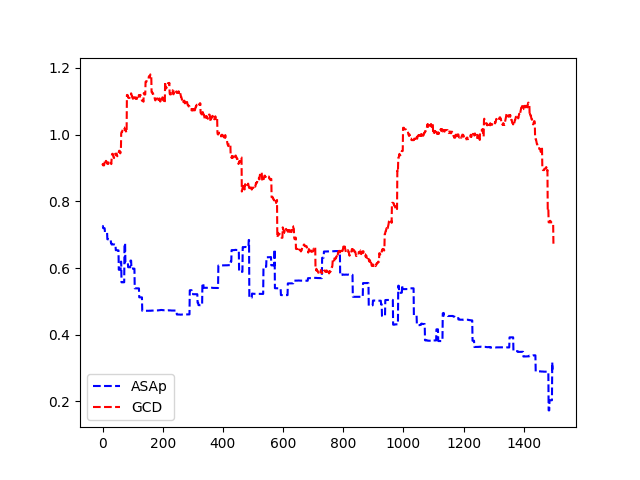}
\caption{KL divergences}
\end{subfigure}
\begin{subfigure}[valign=t]{0.49\textwidth}
\includegraphics[width=\textwidth]{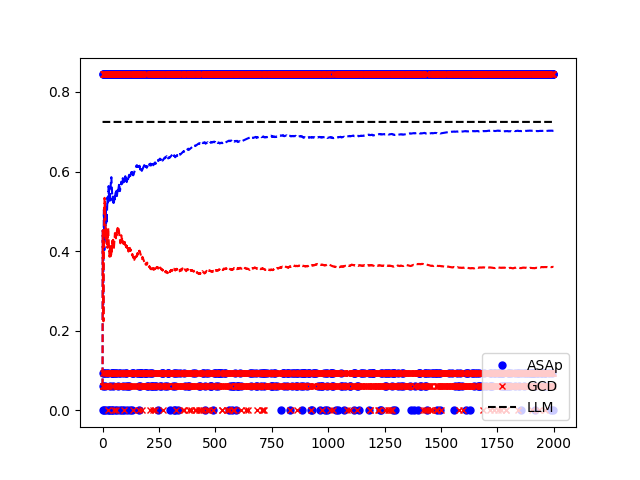}
\caption{Expectations}
\end{subfigure}
\caption{\texttt{INV-BV/find\_inv\_ne\_bvudiv1\_4bit}}
\end{minipage}

\begin{minipage}{0.49\textwidth}
\begin{subfigure}[valign=t]{0.49\textwidth}
\includegraphics[width=\textwidth]{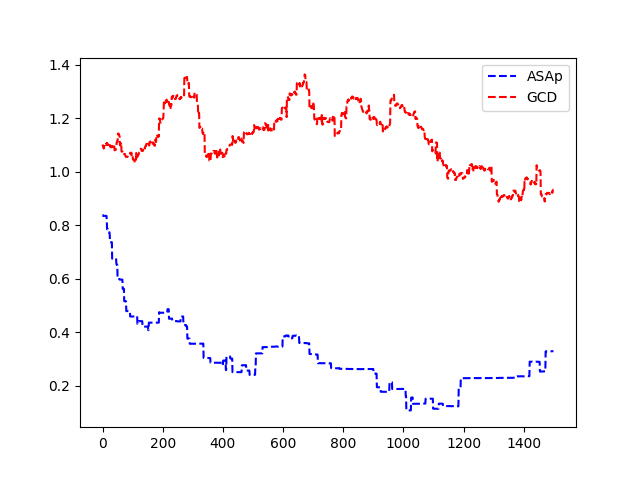}
\caption{KL divergences}
\end{subfigure}
\begin{subfigure}[valign=t]{0.49\textwidth}
\includegraphics[width=\textwidth]{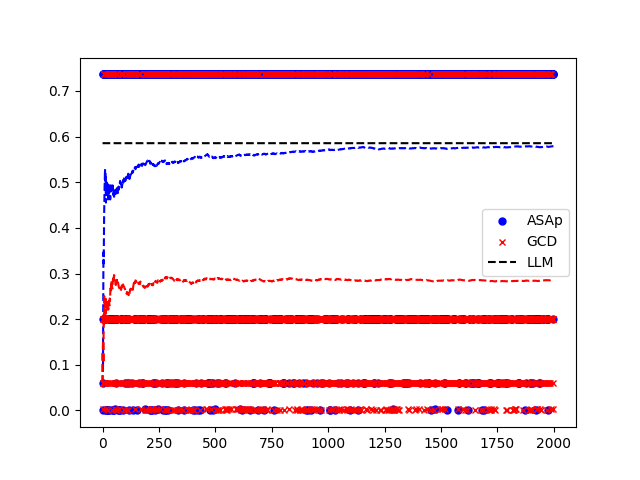}
\caption{Expectations}
\end{subfigure}
\caption{\texttt{INV-BV/find\_inv\_ne\_bvurem0\_4bit}}
\end{minipage}
\hfill
\begin{minipage}{0.49\textwidth}
\begin{subfigure}[valign=t]{0.49\textwidth}
\includegraphics[width=\textwidth]{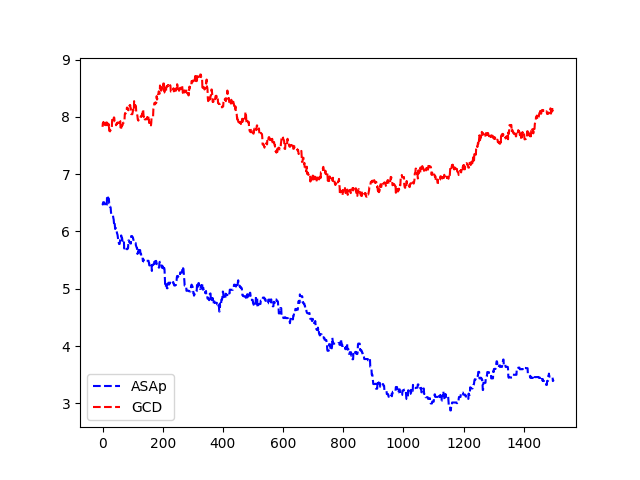}
\caption{KL divergences}
\end{subfigure}
\begin{subfigure}[valign=t]{0.49\textwidth}
\includegraphics[width=\textwidth]{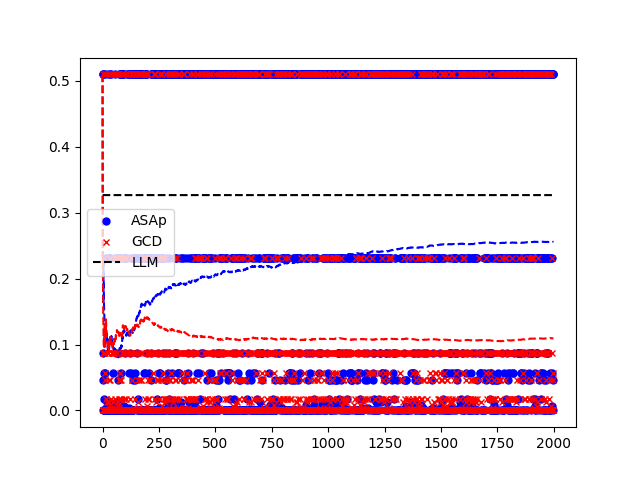}
\caption{Expectations}
\end{subfigure}
\caption{\texttt{INV-BV/find\_inv\_ne\_bvurem1\_4bit}}
\end{minipage}
\end{figure}

\begin{figure}[t]
\begin{minipage}{0.49\textwidth}
\begin{subfigure}[valign=t]{0.49\textwidth}
\includegraphics[width=\textwidth]{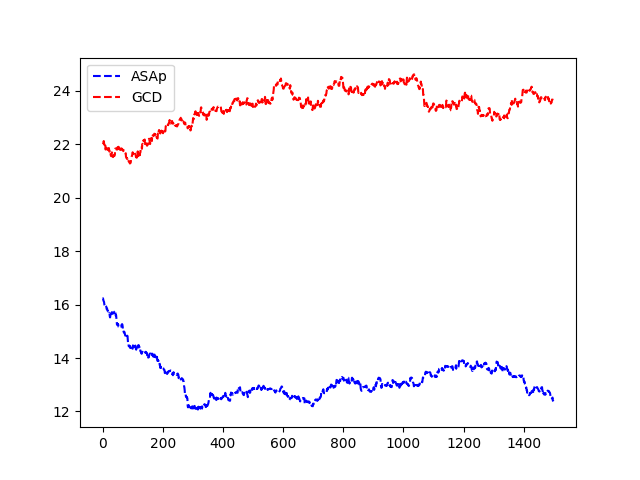}
\caption{KL divergences}
\end{subfigure}
\begin{subfigure}[valign=t]{0.49\textwidth}
\includegraphics[width=\textwidth]{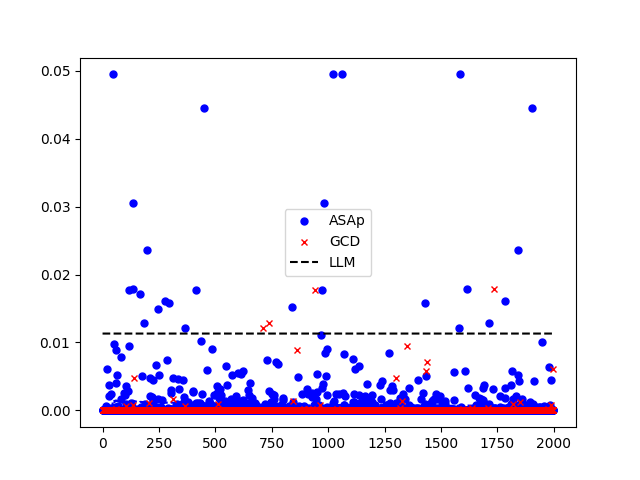}
\caption{Expectations}
\end{subfigure}
\caption{\texttt{CP/CP\_re\_ptb\_460}}
\end{minipage}
\hfill
\begin{minipage}{0.49\textwidth}
\begin{subfigure}[valign=t]{0.49\textwidth}
\includegraphics[width=\textwidth]{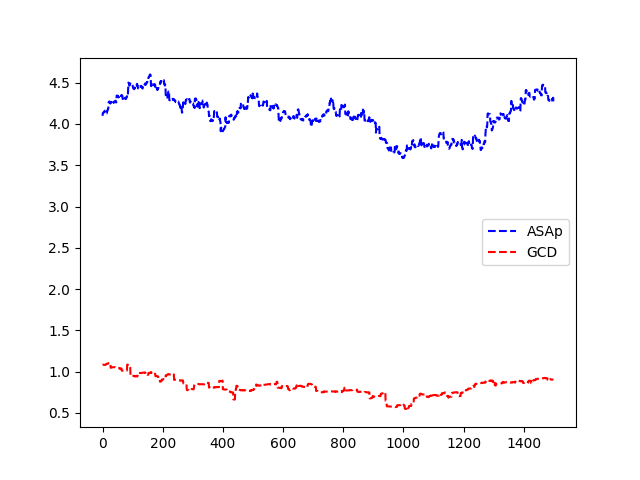}
\caption{KL divergences}
\end{subfigure}
\begin{subfigure}[valign=t]{0.49\textwidth}
\includegraphics[width=\textwidth]{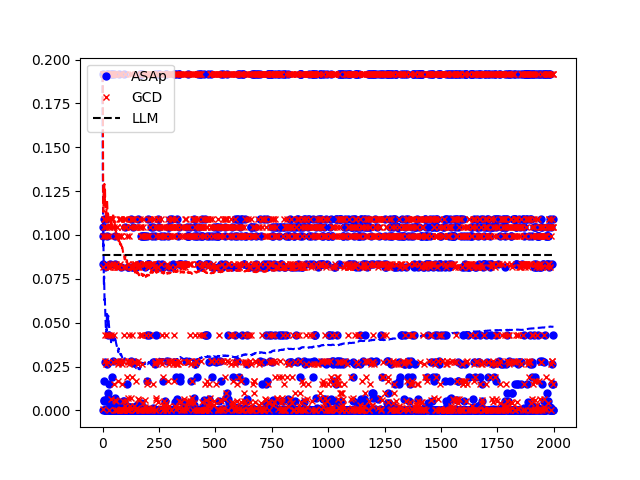}
\caption{Expectations}
\end{subfigure}
\caption{\texttt{CP/CP\_re\_ptb\_482}}
\end{minipage}

\begin{minipage}{0.49\textwidth}
\begin{subfigure}[valign=t]{0.49\textwidth}
\includegraphics[width=\textwidth]{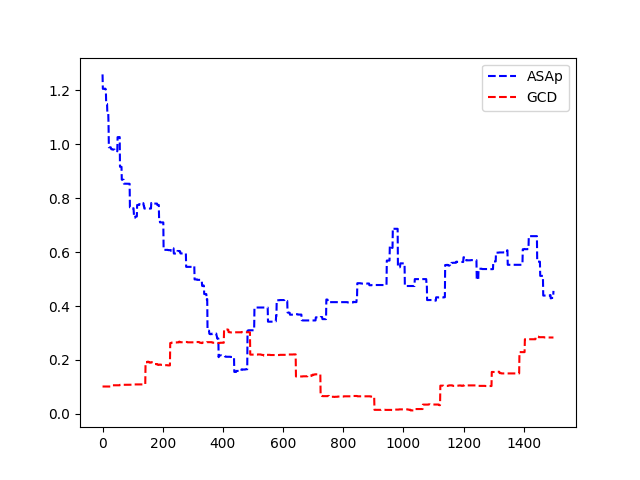}
\caption{KL divergences}
\end{subfigure}
\begin{subfigure}[valign=t]{0.49\textwidth}
\includegraphics[width=\textwidth]{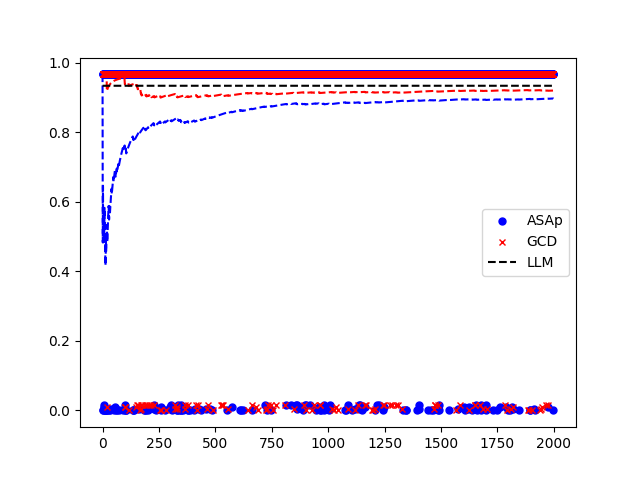}
\caption{Expectations}
\end{subfigure}
\caption{\texttt{CP/CP\_re\_ptb\_486}}
\end{minipage}
\hfill
\begin{minipage}{0.49\textwidth}
\begin{subfigure}[valign=t]{0.49\textwidth}
\includegraphics[width=\textwidth]{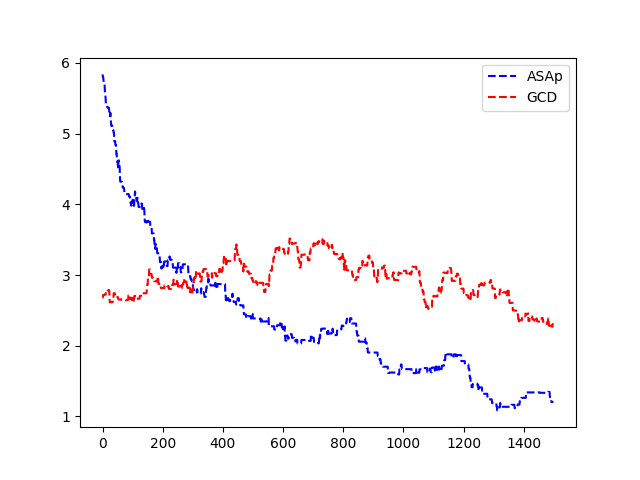}
\caption{KL divergences}
\end{subfigure}
\begin{subfigure}[valign=t]{0.49\textwidth}
\includegraphics[width=\textwidth]{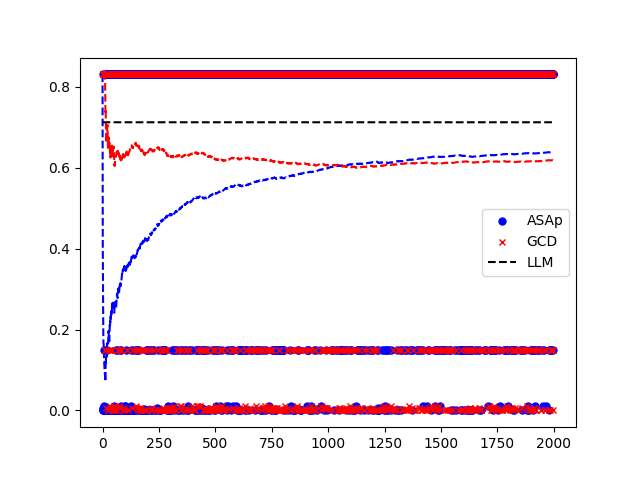}
\caption{Expectations}
\end{subfigure}
\caption{\texttt{CP/CP\_re\_ptb\_605}}
\end{minipage}

\begin{minipage}{0.49\textwidth}
\begin{subfigure}[valign=t]{0.49\textwidth}
\includegraphics[width=\textwidth]{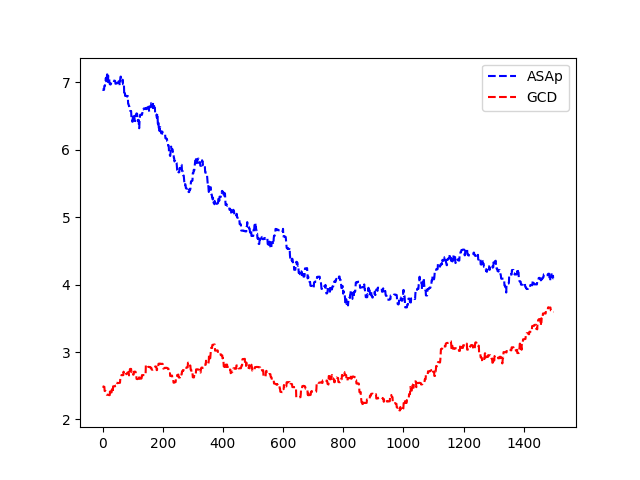}
\caption{KL divergences}
\end{subfigure}
\begin{subfigure}[valign=t]{0.49\textwidth}
\includegraphics[width=\textwidth]{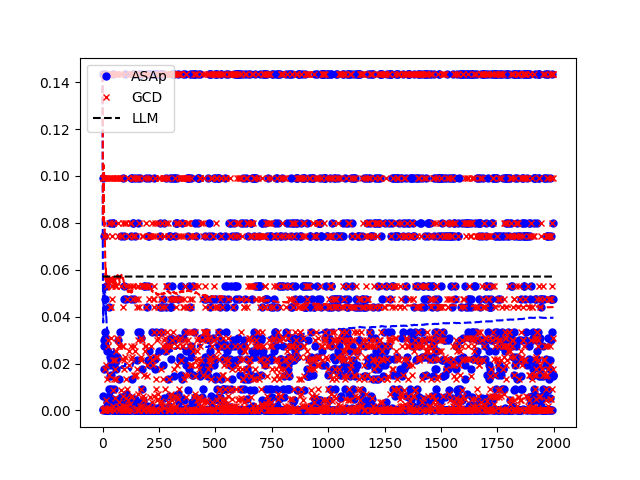}
\caption{Expectations}
\end{subfigure}
\caption{\texttt{CP/CP\_re\_ptb\_1434}}
\end{minipage}
\hfill
\begin{minipage}{0.49\textwidth}
\begin{subfigure}[valign=t]{0.49\textwidth}
\includegraphics[width=\textwidth]{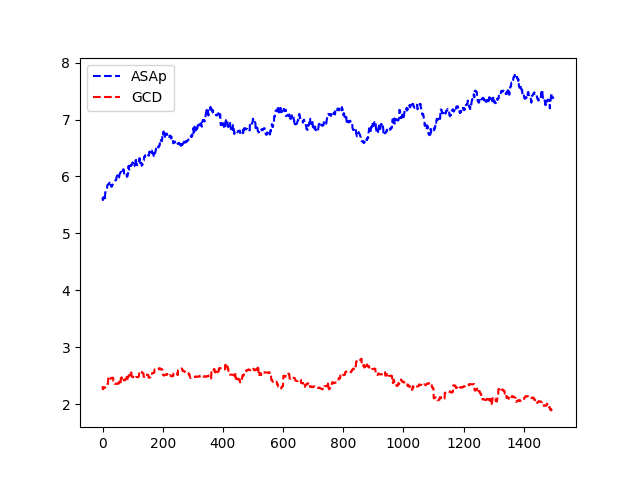}
\caption{KL divergences}
\end{subfigure}
\begin{subfigure}[valign=t]{0.49\textwidth}
\includegraphics[width=\textwidth]{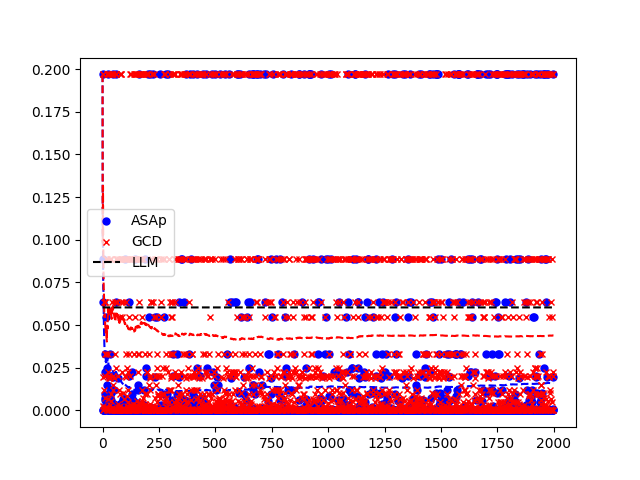}
\caption{Expectations}
\end{subfigure}
\caption{\texttt{CP/CP\_re\_ptb\_1643}}
\end{minipage}
\end{figure}

\begin{table}[h]
    \centering
    \caption{Correctness of solutions for different algorithms.}
    \label{tab:performance}
    \begin{tabular}{llrr}
        \toprule
         & \textbf{Benchmark} & \textbf{Correct ASAP} & \textbf{Correct GCD} \\
        \midrule
        \multirow{16}{*}{SLIA} & phone-3-long & 0 & 0 \\
         & name-combine-2\_short & 171 & \textbf{319} \\
         & name-combine-2-long-repeat & 0 & 0 \\
         & name-combine-4-short & \textbf{20} & 11 \\
         & name-combine-4-long & \textbf{1588} & 12 \\
         & lastname & 285 & \textbf{1526} \\
         & firstname & 1960 & \textbf{1997} \\
         & firstname\_small & 1754 & \textbf{1997} \\
         & reverse-name-long & \textbf{1981} & 1859 \\
         & univ\_1 & \textbf{67} & 40 \\
         & univ\_1\_short & 605 & \textbf{1859} \\
         & univ\_2\_short & 0 & 0 \\
         & dr-name & 357 & \textbf{1654} \\
         & initials\_small & 69 & \textbf{363} \\
         & initials\_long & 540 & \textbf{1584} \\
         & initials\_long-repeat & \textbf{3} & 0 \\\midrule
        \multirow{9}{*}{INV-BV} & find\_inv\_ne\_bvudiv1\_4bit & 0 & 0 \\
        &find\_inv\_bvugt\_bvashr0\_4bit & \textbf{83} & 49 \\
        &find\_inv\_eq\_bvlshr0\_4bit & \textbf{635} & 228 \\
        &find\_inv\_eq\_bvand\_4bit & \textbf{1599} & 1305 \\
        &find\_inv\_bvule\_bvurem0\_4bit & \textbf{1813} & 1710 \\
        &find\_inv\_bvsgt\_bvor\_4bit & \textbf{11} & 10 \\
        &find\_inv\_bvugt\_bvneg\_4bit & \textbf{84} & 36 \\
        &find\_inv\_bvule\_bvurem1\_4bit & 143 & \textbf{227} \\
        \bottomrule
    \end{tabular}
\end{table}

\subsection{Correctness Results for \sygus Tasks}
\label{app:correctness-sygus}
Table~\ref{tab:performance} shows how many samples (out of 2000) yielded correct solutions for each benchmark (bold is better).
The task initials\_long-repeat was only solved using ASAp.

\section{Will ASAp still be more aligned than GCD after fine-tuning?}

\subsection{Experimental setup for fine-tuning}
We adhere to the established QLoRA finetuning pipeline and create task-specific datasets of INV-BV4 and CP for instruction tuning. In line with our paper's methodology, we incorporate in-context examples in the instruction tuning dataset to enhance the models' performance in in-context learning. For each task, we independently finetune Mistral-7B, resulting in two versions of the model (for INV-BV4 and CP). We employ a standard train-validation-test split of 70-10-20\%. Instruction tuning is conducted on the training set, and model selection is based on the lowest validation loss. Key hyperparameters include a learning rate of 2e-4, a warmup ratio of 0.03, a maximum sequence length of 2048, LoRA alpha of 32, LoRA dropout of 0.05, and LoRA rank of 64. The best checkpoints were at 328 and 536 steps for INV-BV and CP, respectively.
\subsection{Additional Results}

\paragraph{No significant differences in convergence rates post fine-tuning.} 

In Section~\ref{sec:eval}, we evaluate ASAp and GCD on the base model Mistral-7B. A natural extension of this evaluation is determining whether ASAp retains its advantages over GCD after fine-tuning the base model on task-specific datasets, which optimizes the LLM for higher grammatical accuracy from the start.

In our fine-tuning step, we want to teach the LLM to assign higher probabilities to grammatical outputs for the specific task DSL. We randomly selected two INV-BV problems (\texttt{find\_inv\_bvsge\_bvneg\_4bit} and \texttt{find\_inv\_bvsgt\_bvor\_4bit} for INV-BV) and four CP problems (\texttt{CP\_re\_ptb\_215}, \texttt{CP\_re\_ptb\_434}, \texttt{CP\_re\_ptb\_1627} and \texttt{CP\_re\_ptb\_1643} for CP) from the test set, and instrction tuned input-output pairs of prompt and output programs in the training set for the base model Mistral-7B. We obtained two fine-tuned models, one for INV-BV and one for CP.

We tested GCD and ASAp on the finetuned Mistral-7B on the randomly left-out problems and checked the convergence rates of the KL-divergence. The results from finetuned Mistral-7B did not show significant differences in terms of convergence compared to the base Mistral-7B. As done in Section~\ref{sec:eval}, we computed the expectation for each benchmark obtained via GCD and ASAp after 2,000 iterations and compared it against the target expectation $\probgrammarpg{\prob}{\grammar}$ of GAD. The sum of least squares difference between expectations computed by GCD and the expectations of $\probgrammarpg{\prob}{\grammar}$ are 0.677 (INV-BV4), 0.278 (CP), while ASAp achieved lower errors: 0.051 (INV-BV4), 0.201 (CP), indicating that ASAp more closely aligned with the exact GAD expectations. We did not include SLIA as we did not have sufficient data for further fine-tuning.

%% file: main.bib
@inproceedings{kumar2022gradient,
  title={Gradient-based constrained sampling from language models},
  author={Kumar, Sachin and Paria, Biswajit and Tsvetkov, Yulia},
  booktitle={Proceedings of the 2022 Conference on Empirical Methods in Natural Language Processing},
  pages={2251--2277},
  year={2022}
}

@inproceedings{qin2022cold,
  title={COLD Decoding: Energy-based Constrained Text Generation with Langevin Dynamics},
  author={Qin, Lianhui and Welleck, Sean and Khashabi, Daniel and Choi, Yejin},
  booktitle={Advances in Neural Information Processing Systems},
  year={2022}
}

@misc{wang2023grammar,
      title={Grammar Prompting for Domain-Specific Language Generation with Large Language Models}, 
      author={Bailin Wang and Zi Wang and Xuezhi Wang and Yuan Cao and Rif A. Saurous and Yoon Kim},
      year={2023},
      eprint={2305.19234},
      archivePrefix={arXiv},
      primaryClass={cs.CL}
}

@inproceedings{geng2024grammarconstrained,
	title        = {Grammar-Constrained Decoding for Structured {NLP} Tasks without Finetuning},
	author       = {Geng, Saibo  and Josifoski, Martin  and Peyrard, Maxime  and West, Robert},
	year         = 2023,
	month        = dec,
	booktitle    = {Proceedings of the 2023 Conference on Empirical Methods in Natural Language Processing},
	publisher    = {Association for Computational Linguistics},
	address      = {Singapore},
	url          = {https://aclanthology.org/2023.emnlp-main.674},
	editor       = {Bouamor, Houda  and Pino, Juan  and Bali, Kalika}
}

@article{willard2023efficient,
  title={Efficient Guided Generation for Large Language Models},
  author={Willard, Brandon T and Louf, R{\'e}mi},
  journal={arXiv e-prints},
  pages={arXiv--2307},
  year={2023}
}

@misc{alur2019syguscomp,
      title={SyGuS-Comp 2018: Results and Analysis}, 
      author={Rajeev Alur and Dana Fisman and Saswat Padhi and Rishabh Singh and Abhishek Udupa},
      year={2019},
      eprint={1904.07146},
      archivePrefix={arXiv},
      primaryClass={cs.PL}
}

@misc{Geng2023,
  author = {Geng, Saibo  and Josifoski, Martin  and Peyrard, Maxime  and West, Robert},
  title = {Transformers-{CFG}},
  year = {2023},
  publisher = {GitHub},
  journal = {GitHub repository},
  howpublished = {\url{https://github.com/epfl-dlab/transformers-CFG}}
}

@inproceedings{AgrawalKGLR23,
  author       = {Lakshya A Agrawal and
                  Aditya Kanade and
                  Navin Goyal and
                  Shuvendu K. Lahiri and
                  Sriram K. Rajamani},
  editor       = {Alice Oh and
                  Tristan Naumann and
                  Amir Globerson and
                  Kate Saenko and
                  Moritz Hardt and
                  Sergey Levine},
  title        = {Monitor-Guided Decoding of Code LMs with Static Analysis of Repository
                  Context},
  booktitle    = {Advances in Neural Information Processing Systems 36: Annual Conference
                  on Neural Information Processing Systems 2023, NeurIPS 2023, New Orleans,
                  LA, USA, December 10 - 16, 2023},
  year         = {2023},
  url          = {http://papers.nips.cc/paper\_files/paper/2023/hash/662b1774ba8845fc1fa3d1fc0177ceeb-Abstract-Conference.html},
  bibsource    = {dblp computer science bibliography, https://dblp.org}
}

@article{lmql,
author = {Beurer-Kellner, Luca and Fischer, Marc and Vechev, Martin},
title = {Prompting Is Programming: A Query Language for Large Language Models},
year = {2023},
issue_date = {June 2023},
publisher = {Association for Computing Machinery},
address = {New York, NY, USA},
volume = {7},
number = {PLDI},
url = {https://doi.org/10.1145/3591300},
doi = {10.1145/3591300},
journal = {Proc. ACM Program. Lang.},
month = {jun},
articleno = {186},
numpages = {24}
}

@inproceedings{picard,
    title = "{PICARD}: Parsing Incrementally for Constrained Auto-Regressive Decoding from Language Models",
    author = "Scholak, Torsten  and
      Schucher, Nathan  and
      Bahdanau, Dzmitry",
    editor = "Moens, Marie-Francine  and
      Huang, Xuanjing  and
      Specia, Lucia  and
      Yih, Scott Wen-tau",
    booktitle = "Proceedings of the 2021 Conference on Empirical Methods in Natural Language Processing",
    month = nov,
    year = "2021",
    address = "Online and Punta Cana, Dominican Republic",
    publisher = "Association for Computational Linguistics",
    url = "https://aclanthology.org/2021.emnlp-main.779",
    doi = "10.18653/v1/2021.emnlp-main.779",
    pages = "9895--9901",
}

@inproceedings{llm_sampling_renda_hopkins_2023,
title={Can {LLM}s Generate Random Numbers? Evaluating {LLM} Sampling in Controlled Domains},
author={Renda, Alex and Hopkins, Aspen K. and Carbin, Michael},
booktitle={ICML 2023 Workshop: Sampling and Optimization in Discrete Space},
year={2023},
url={http://people.csail.mit.edu/renda/llm-sampling-paper},
}

@misc{huang2024large,
      title={Large Language Models Based Fuzzing Techniques: A Survey}, 
      author={Linghan Huang and Peizhou Zhao and Huaming Chen and Lei Ma},
      year={2024},
      eprint={2402.00350},
      archivePrefix={arXiv},
      primaryClass={cs.SE}
}

@article{jiang2023mistral,
  title={Mistral 7B},
  author={Jiang, Albert Q and Sablayrolles, Alexandre and Mensch, Arthur and Bamford, Chris and Chaplot, Devendra Singh and Casas, Diego de las and Bressand, Florian and Lengyel, Gianna and Lample, Guillaume and Saulnier, Lucile and others},
  journal={arXiv preprint arXiv:2310.06825},
  year={2023}
}

@article{poesia2022synchromesh,
  title={Synchromesh: Reliable code generation from pre-trained language models},
  author={Poesia, Gabriel and Polozov, Oleksandr and Le, Vu and Tiwari, Ashish and Soares, Gustavo and Meek, Christopher and Gulwani, Sumit},
  journal={arXiv preprint arXiv:2201.11227},
  year={2022}
}

@article{dong2022codepad,
  title={Codepad: Sequence-based code generation with pushdown automaton},
  author={Dong, Yihong and Jiang, Xue and Liu, Yuchen and Li, Ge and Jin, Zhi},
  journal={arXiv preprint arXiv:2211.00818},
  year={2022}
}

@article{melcer2024constrained,
  title={Constrained Decoding for Code Language Models via Efficient Left and Right Quotienting of Context-Sensitive Grammars},
  author={Melcer, Daniel and Fulton, Nathan and Gouda, Sanjay Krishna and Qian, Haifeng},
  journal={arXiv preprint arXiv:2402.17988},
  year={2024}
}

@article{ugare2024improving,
  title={Improving llm code generation with grammar augmentation},
  author={Ugare, Shubham and Suresh, Tarun and Kang, Hangoo and Misailovic, Sasa and Singh, Gagandeep},
  journal={arXiv preprint arXiv:2403.01632},
  year={2024}
}

@inproceedings{li2024guiding,
  title={Guiding enumerative program synthesis with large language models},
  author={Li, Yixuan and Parsert, Julian and Polgreen, Elizabeth},
  booktitle={International Conference on Computer Aided Verification},
  pages={280--301},
  year={2024},
  organization={Springer}
}

@article{shin2021constrained,
  title={Constrained language models yield few-shot semantic parsers},
  author={Shin, Richard and Lin, Christopher H and Thomson, Sam and Chen, Charles and Roy, Subhro and Platanios, Emmanouil Antonios and Pauls, Adam and Klein, Dan and Eisner, Jason and Van Durme, Benjamin},
  journal={arXiv preprint arXiv:2104.08768},
  year={2021}
}

@article{hokamp2017lexically,
  title={Lexically constrained decoding for sequence generation using grid beam search},
  author={Hokamp, Chris and Liu, Qun},
  journal={arXiv preprint arXiv:1704.07138},
  year={2017}
}

@article{anderson2016guided,
  title={Guided open vocabulary image captioning with constrained beam search},
  author={Anderson, Peter and Fernando, Basura and Johnson, Mark and Gould, Stephen},
  journal={arXiv preprint arXiv:1612.00576},
  year={2016}
}

@article{post2018fast,
  title={Fast lexically constrained decoding with dynamic beam allocation for neural machine translation},
  author={Post, Matt and Vilar, David},
  journal={arXiv preprint arXiv:1804.06609},
  year={2018}
}

@inproceedings{hu2019improved,
  title={Improved lexically constrained decoding for translation and monolingual rewriting},
  author={Hu, J Edward and Khayrallah, Huda and Culkin, Ryan and Xia, Patrick and Chen, Tongfei and Post, Matt and Van Durme, Benjamin},
  booktitle={Proceedings of the 2019 Conference of the North American Chapter of the Association for Computational Linguistics: Human Language Technologies, Volume 1 (Long and Short Papers)},
  pages={839--850},
  year={2019}
}

@inproceedings{lu-etal-2021-neurologic,
    title = "{N}euro{L}ogic Decoding: (Un)supervised Neural Text Generation with Predicate Logic Constraints",
    author = "Lu, Ximing  and
      West, Peter  and
      Zellers, Rowan  and
      Le Bras, Ronan  and
      Bhagavatula, Chandra  and
      Choi, Yejin",
    editor = "Toutanova, Kristina  and
      Rumshisky, Anna  and
      Zettlemoyer, Luke  and
      Hakkani-Tur, Dilek  and
      Beltagy, Iz  and
      Bethard, Steven  and
      Cotterell, Ryan  and
      Chakraborty, Tanmoy  and
      Zhou, Yichao",
    booktitle = "Proceedings of the 2021 Conference of the North American Chapter of the Association for Computational Linguistics: Human Language Technologies",
    month = jun,
    year = "2021",
    address = "Online",
    publisher = "Association for Computational Linguistics",
    url = "https://aclanthology.org/2021.naacl-main.339",
    doi = "10.18653/v1/2021.naacl-main.339",
    pages = "4288--4299",
}

@inproceedings{lu-etal-2022-neurologic,
    title = "{N}euro{L}ogic A*esque Decoding: Constrained Text Generation with Lookahead Heuristics",
    author = "Lu, Ximing  and
      Welleck, Sean  and
      West, Peter  and
      Jiang, Liwei  and
      Kasai, Jungo  and
      Khashabi, Daniel  and
      Le Bras, Ronan  and
      Qin, Lianhui  and
      Yu, Youngjae  and
      Zellers, Rowan  and
      Smith, Noah A.  and
      Choi, Yejin",
    editor = "Carpuat, Marine  and
      de Marneffe, Marie-Catherine  and
      Meza Ruiz, Ivan Vladimir",
    booktitle = "Proceedings of the 2022 Conference of the North American Chapter of the Association for Computational Linguistics: Human Language Technologies",
    month = jul,
    year = "2022",
    address = "Seattle, United States",
    publisher = "Association for Computational Linguistics",
    url = "https://aclanthology.org/2022.naacl-main.57",
    doi = "10.18653/v1/2022.naacl-main.57",
    pages = "780--799",
}

@article{li2022diffusion,
  title={Diffusion-lm improves controllable text generation},
  author={Li, Xiang and Thickstun, John and Gulrajani, Ishaan and Liang, Percy S and Hashimoto, Tatsunori B},
  journal={Advances in Neural Information Processing Systems},
  volume={35},
  pages={4328--4343},
  year={2022}
}

@article{amini2024structured,
  title={Structured voronoi sampling},
  author={Amini, Afra and Du, Li and Cotterell, Ryan},
  journal={Advances in Neural Information Processing Systems},
  volume={36},
  year={2024}
}

@article{stengel2023zero,
  title={Zero and few-shot semantic parsing with ambiguous inputs},
  author={Stengel-Eskin, Elias and Rawlins, Kyle and Van Durme, Benjamin},
  journal={arXiv preprint arXiv:2306.00824},
  year={2023}
}
